\documentclass[twoside,11pt]{article}
\pdfoutput=1 

\usepackage[nohyperref]{automl2019}

\usepackage{amsfonts}
\usepackage{amssymb}
\usepackage{amsmath}
\usepackage{geometry}
\usepackage{graphicx}
\usepackage{physics} 
\usepackage{wrapfig}
\usepackage[protrusion=true,tracking=true,kerning=true,expansion,final]{microtype}      
\usepackage{xcolor} 
\usepackage{booktabs} 
\usepackage{algorithmicx}
\usepackage{algcompatible}
\usepackage[section]{algorithm}
\usepackage{pdflscape}
\usepackage{subcaption}

\usepackage{varioref} 
\usepackage[pagebackref]{hyperref}
\usepackage[capitalise]{cleveref} 
\crefname{appsec}{appendix}{appendices}
\Crefname{appsec}{Appendix}{Appendices}

\usepackage{url}

\renewcommand{\cite}[1]{\citep{#1}}
\definecolor{mydarkblue}{rgb}{0,0.08,0.45}
\hypersetup{ %
	pdftitle={Towards  Learning of Filter-Level Heterogeneous Compression of Convolutional Neural Networks}, 
	pdfauthor={Yochai Zur, Chaim Baskin, Evgenii Zheltonozhskii, Brian Chmiel, Itay Evron and Alex M.\ Bronstein},
	pdfsubject={},
	pdfkeywords={},
	pdfborder=0 0 0,
	pdfpagemode=UseNone,
	colorlinks=true,
	allbordercolors={1 1 1},
	linkcolor=mydarkblue,
	citecolor=mydarkblue,
	filecolor=mydarkblue,
	urlcolor=mydarkblue,
	pdfview=FitH}

\renewcommand*{\backref}[1]{} 
\renewcommand*{\backrefalt}[4]{%
	\ifcase #1 %
	\or
	(cited on p. #2)%
	\else
	(cited on pp. #2)%
	\fi
}

\vbadness=\maxdimen

\usepackage{chngcntr}
\usepackage{apptools}
\AtAppendix{\counterwithin{figure}{section}}
\AtAppendix{\counterwithin{table}{section}}

\jmlrheading{Yochai Zur, Chaim Baskin, Evgenii Zheltonozhskii, Brian Chmiel, Itay Evron, Alex M.\ Bronstein and Avi Mendelson}

\ShortHeadings{Towards  Learning of Filter-Level Heterogeneous Compression of CNNs}{Zur, Baskin, Zheltonozhskii, Chmiel, Evron, Bronstein and Mendelson} 
\firstpageno{1}

\usepackage{bm}

\newcommand{\ie}{\textit{i}.\textit{e}., }
\newcommand{\eg}{\textit{e}.\textit{g}., }

\newcommand{\etal}{\textit{et al}.\ }

\newcommand{\bb}[1]{\bm{#1}}

\def\layerOpsSet{T_\ell}

\def\nFiltersLayer{C_\ell}
\def\configIndices{1,2,\dots,L}
\newcommand{\configSubset}[1]{S_{#1}}
\def\networkConfig{\bb{a}}
\newcommand{\homogeneousConfig}[1]{a_{h_#1}}
\def\networkComplexity{z_{\configIndices}}
\newcommand{\homogeneousComplexity}[1]{z_{h_#1}}
\def\networkWeights{\bb{\omega}}
\def\configWeights{\networkWeights_{\configIndices}}
\def\trainSet{T}
\def\searchSet{T_{\alpha}}
\def\trainWeightsSet{T_{\networkWeights}}
\def\trainWeightsEpochs{t_{\networkWeights}}
\def\trainWeightsInterval{k_{\networkWeights}}
\def\probConfig{p(\bb{a} | \bb{\alpha})}
\def\Loss{\mathcal{L}}
\def\prob{\hat{\alpha}}
\def\lossConfig{\Loss (\networkConfig;\networkWeights)}

\newcommand{\expectedConfig}[1]{\mathcal{A}^{\prob^{#1}}} 

\newcommand{\ceLossGeneric}[1]{\Loss_{\text{acc}} \left( \networkConfig;\networkWeights_{#1} \right)}
\def\ceLoss{\ceLossGeneric{}}

\newcommand{\ceLossInterpolation}[1]{\Loss_{\text{acc}}^{h_1,h_2} \left( #1;\networkWeights \right)}
\def\complexitySigma{\sigma \left( \cdot \right)}
\def\complexityLoss{\Loss_{\text{com}} (\networkConfig)}
\def\J{J(\bb{\alpha} ;\networkWeights)}
\def\configsSumOperator{\sum_{\bb{a}}} 
\def\configsSum{\configsSumOperator \probConfig \cdot \lossConfig}
\newcommand{\opAlpha}[1]{\alpha_{\ell #1}}
\def\opAlphaSingle{\alpha_{\ell}}
\newcommand{\opSample}[1]{a_{\ell #1}}
\def\opSampleSingle{a_{\ell}}
\newcommand{\opProb}[1]{\hat{\alpha}_{\ell #1}}
\def\opProbSingle{\hat{\alpha}_{\ell}}
\newcommand{\opDerivative}[1]{g_{\opAlpha{#1}}}

\newcommand{\opDerivativeEstimator}[1]{\hat{g}_{\opAlpha{#1}}}
\newcommand{\opProbDetailed}[1]{\frac{\exp \left\{ \opAlpha{#1} \right\}}{\sum_{j=1}^{\abs{\layerOpsSet}} \exp \left\{ \opAlpha{j} \right\}}}
\newcommand{\layerConfig}[1]{a_{{#1}}}
\newcommand{\layerRV}[1]{\bb{A}_{#1}}
\newcommand{\prodLayersProb}[1]{\prod_{{#1}=1}^L \Pr \left( \layerRV{#1} = \layerConfig{#1} \right)}
\def\layerConfigDetailed{\layerConfig{\ell} = (\opSample{1},\dots,\opSample{\abs{T_\ell}})}
\def\layerConfigSingleDetailed{\layerConfig{\ell} = (\opSampleSingle)}
\def\bitWeights{b_\mathrm{w}}
\def\bitActivations{b_\mathrm{a}}
\def\bitwidthTuple{\left( \bitWeights,\bitActivations \right)}

\newcommand{\layerOpsProd}[1]{\prod_{#1=1}^{\abs{\layerOpsSet}}}
\def\layerOpsProdK{\layerOpsProd{k}}
\newcommand{\layerOpsSum}[1]{\sum_{#1=1}^{\abs{\layerOpsSet}}}
\def\layerOpsSumK{\layerOpsSum{k}}
\def\multinomialFrac{\frac{\nFiltersLayer !}{\layerOpsProdK \opSample{k}}}
\def\opSampleNumeratorMultinomial{\left( \exp \left\{ \opAlpha{k} \cdot \opSample{k} \right\} \right)}
\def\opSampleNumeratorMultinomialSum{\exp \left\{ \layerOpsSumK \opAlpha{k} \cdot \opSample{k} \right\}}
\def\opSampleDenominatorMultinomial{\sum_{j=1}^{\abs{\layerOpsSet}} \exp \left\{ \opAlpha{j} \right\}}
\def\opSampleDenominatorMultinomialWithPower{\left( \opSampleDenominatorMultinomial \right) ^{\nFiltersLayer}}
\def\multinomialConfigProbFrac{\frac{\opSampleNumeratorMultinomialSum}{\opSampleDenominatorMultinomialWithPower}}
\def\multinomialConfigProb{\multinomialFrac \cdot \multinomialConfigProbFrac}
\def\layerConfigProb{\Pr \qty( \layerRV{\ell} = \layerConfig{\ell} )}
\def\opAlphaDerivative{\pdv{\opAlpha{t}}}
\def\opAlphaSingleDerivative{\pdv{\opAlphaSingle}}
\newcommand{\opSampleConfig}[1]{\opSample{t}^{#1}}
\newcommand{\multinomialConfigDerivativeDifferencePart}[1]{\left( \opSampleConfig{#1} - \nFiltersLayer \cdot \opProb{t} \right)}
\def\multinomialConfigDerivative{\multinomialConfigDerivativeDifferencePart{} \cdot \layerConfigProb}
\newcommand{\prodLayersProbLeaveOut}[1]{\prod_{#1 \neq \ell} \Pr \left( \layerRV{#1} = \layerConfig{#1} \right)}
\newcommand{\multinomialConfigProbDerivative}[1]{\multinomialConfigDerivativeDifferencePart{#1} \cdot \probConfig}
\def\multinomialLossEVderivative{\configsSumOperator \lossConfig \cdot \multinomialConfigProbDerivative{}}
\def\nTrialsBinomial{\left( \nFiltersLayer-1 \right)}
\def\opProbBinomialDetailedNumerator{\exp \left\{ \opAlphaSingle \right\}}
\def\opProbBinomialDetailedNumeratorWithPower{\exp \left\{ \opAlphaSingle \cdot \opSampleSingle \right\}}
\def\opProbBinomialDetailedDenominator{\exp \left\{ \opAlphaSingle \right\} + 1}
\def\opProbBinomialDetailedDenominatorWithPower{\qty\big( \opProbBinomialDetailedDenominator ) ^{\nTrialsBinomial}}
\def\opProbBinomialDetailed{\frac{\exp \left\{ \opAlphaSingle \right\}}{\opProbBinomialDetailedDenominator}}
\def\binomialFrac{\frac{\nTrialsBinomial !}{\opSampleSingle ! \cdot \left( \nFiltersLayer-1 - \opSampleSingle \right) !}}
\def\opProbBinomialDetailedWithPower{\left( \opProbBinomialDetailed \right) ^{\opSampleSingle}}
\def\binomialComplementPower{\nTrialsBinomial - \opSampleSingle}
\def\binomialConfigProbFrac{\frac{\opProbBinomialDetailedNumeratorWithPower}{\opProbBinomialDetailedDenominatorWithPower}}
\def\binomialProb{\frac{\opProbBinomialDetailedNumerator}{\opProbBinomialDetailedDenominator}}
\def\binomialConfigProb{\binomialFrac \cdot \binomialConfigProbFrac}
\newcommand{\opSampleSingleConfig}[1]{\opSampleSingle^{#1}}
\newcommand{\binomialConfigDerivativeDifferencePart}[1]{\left( \opSampleSingleConfig{#1} - \nTrialsBinomial \cdot \opProbSingle \right)}
\def\binomialConfigDerivative{\binomialConfigDerivativeDifferencePart{} \cdot \layerConfigProb}
\newcommand{\binomialConfigProbDerivative}[1]{\binomialConfigDerivativeDifferencePart{#1} \cdot \probConfig}
\def\binomialLossEVderivative{\configsSumOperator \lossConfig \cdot \binomialConfigProbDerivative{}}

\def\tdu#1#2#3{#1{}_{{#2}}{}^{{#3}}}
\def\oper{{t}}

\def\Loss{{\mathcal{L}}}

\def\bops{{\mathcal{B}}}

\def\idx{{a}}
\def\vidx{{\bm{\idx}}} 

\def\bopsLoss{\complexityLoss}

\def\operations{T}

\newcommand{\multinomial}{\text{Multinomial}}
\newcommand{\binomial}{\text{Binomial}}
\newcommand{\round}{\text{round}}
\begin{document}

\title{Towards  Learning of Filter-Level Heterogeneous Compression of Convolutional Neural Networks}
\author{\name Yochai Zur \thanks{equal contributors} \email yochaiz@cs.technion.ac.il  \\ 
	\name Chaim Baskin \footnotemark[1] \email chaimbaskin@cs.technion.ac.il\\
	\name Evgenii Zheltonozhskii \email evgeniizh@campus.technion.ac.il\\
	\name Brian Chmiel \email brianch@campus.technion.ac.il\\
	\name Itay Evron \email evron.itay@gmail.com \\ 
	\name Alex M.\ Bronstein  \email bron@cs.technion.ac.il\\
	\name Avi Mendelson  \email avi.mendelson@tce.technion.ac.il\\
\addr Department of Computer Science, Technion, Haifa, Israel}

\maketitle

\begin{abstract}
Recently, deep learning has become a \emph{de facto} standard in machine learning with convolutional neural networks (CNNs) demonstrating spectacular success on a wide variety of tasks. However, CNNs are typically very demanding computationally at inference time. One of the ways to alleviate this burden on certain hardware platforms is \emph{quantization} relying on the use of low-precision arithmetic representation for the weights and the activations. Another popular method 
is the pruning of the number of filters in each layer. While mainstream deep learning methods train the neural networks weights while keeping the network architecture fixed, 
the emerging neural architecture search (NAS) techniques make the latter also amenable to training.
In this paper, we formulate optimal arithmetic bit length allocation and neural network pruning as a NAS problem, searching for the 
configurations
satisfying a computational complexity budget while maximizing the 
accuracy. We use a differentiable search method based on the continuous relaxation of the 
search space proposed by \citet{liu2018darts}. We show, by grid search, that heterogeneous quantized networks suffer from a high variance which renders the benefit of the search questionable. For pruning, improvement over homogeneous cases is possible, but it is still challenging to find those configurations with the proposed method. The code is publicly available at \href{https://github.com/yochaiz/Slimmable}{https://github.com/yochaiz/Slimmable} and \href{https://github.com/yochaiz/darts-UNIQ}{https://github.com/yochaiz/darts-UNIQ}.

\end{abstract}

\section{Introduction}
\label{sec:intro}

Convolutional neural networks (CNNs) have become a main solution for computer vision tasks. However, high computation requirements complicate their usage in low-power systems. 
Recently, the machine learning approaches outperformed humans in design of CNNs \cite{zoph2018learning,real2018regularized,macko2019improving,ghiasi2019nasfpn} and allowed to optimize complexity \cite{cai2018proxylessnas,guo2019single} or runtime \cite{tan2018mnasnet,chu2019multi,stamoulis2019single}.  Moreover, using gradient-based methods 
\cite{liu2018darts,noy2019asap} reduces search time to a couple of GPU-days.

We focus on two aspects of complexity reduction: quantization and pruning.
Some recent works demonstrated that using $16-$ or even $8-$bit representations do not harm accuracy of NNs \cite{gupta2015deep,jacob2018quantization, lee2018quantization}. To reduce both runtime and power consumption, researches investigated further reduction of bitwidth, up to a single bit \cite{hubara2016binarized, bethge2018training, ding2019regularizing}, which is impossible with naive techniques. 

Alternatively, one could
use fewer convolutional filters by proportionally reducing their number \cite{sandler2018mobilenetv2} or by pruning 
insignificant ones \cite{lecun1990optimal,li2016pruning}. 
Recently, 
slimmable networks \cite{yu2018slimmable} offered a 
method to simultaneously 
train multiple instances of a  
CNN with different filter count. 

Both methods compress the network with some parameter $\alpha$ representing the bitwidth or the percentage of filters pruned. \emph{Homogeneous} configurations use same value of $\alpha$ 
along the network,
\eg same bitwidth of parameters
in each layer.
Many quantization works employ simple \emph{heterogeneous} configurations with different bitwidth 
in first or last layer \cite{zhou2016dorefa, ZhangYangYeECCV2018,hoffer2018fix,mckinstry2018discovering, choi2018bridging}. Some works 
studied
a \emph{layer-wise quantization granularity} \cite{louizos2017bayesian,lacey2018stochastic}. Heterogeneous configurations are often found in pruning too. 

Recent works studied quantization with filter-wise quantization granularity \cite{wu2018mixed,elthakeb2018releq,chen2018joint,lou2019autoqb,guo2019single} and pruning \cite{liu2019metapruning,yu2019network} using NAS techniques.
In this paper we study the opportunities of compression of the network with \emph{filter-wise granularity}. 

\paragraph{Contribution}
The main contribution of this paper are follows: study of variance of compressed networks; arithmetic compression on filter level; application of differentiable NAS to those problems.

The rest of the paper is organized as following: \cref{sec:method} introduces a general method, \cref{sec:quant_exp,sec:pruning_exp} describe experiments performed, and \cref{sec:conclusions} concludes the paper.

\section{Method}
\label{sec:method}

\label{sec:diffsearch}
\paragraph{Differentiable search method}

For CNN with $L$ convolutional layers with $\nFiltersLayer$ filters in layer $\ell=1,\dots,L$, 
let $\layerOpsSet$ denote the set of 
compression operations that can be applied to one filter. 
Our goal is to find an optimal 
assignment of the said operations to each filter.

To do that, we aim to learn a probability of each operation to be chosen, similarly to \citet{liu2018darts}. 
Referring to the operations in $\layerOpsSet$ simply by their index, $i=1,\dots,\abs{\layerOpsSet}$, we assign
to each operation $i$ in layer $\ell$  a 
parameter $\opAlpha{i}$, and denote by $\opProb{i} = f\qty(\opAlpha{i})$ the probability to choose this particular operation.
We also denote 
by $\bb{\alpha}_\ell = (\opAlpha{1},\dots, \opAlpha{|T_\ell| })^{\mathrm{T}}$ 
the vector of parameters in a given layer $\ell$, and by the pseudomatrix $\bb{\alpha} = (\bb{\alpha}_1,\dots, \bb{\alpha}_L)$ the corresponding parameters of the entire network.

To reduce the search space size, 
instead of assigning operations to each filter independently we do it on layer level. Let $\opSample{i}$ denote the number of filters in the layer $\ell$ to which the operation $i$ is applied. We refer to the vector $\bb{a}_{\ell} = (\opSample{1},\dots,\opSample{\abs{T_\ell}})^\mathrm{T}$  with the elements summing to $\nFiltersLayer$ defining the choice of the operations in layer $\ell$ as to the \emph{configuration} of the layer. We denote the configuration of the entire network by the pseudo-matrix $\bb{a}$.

Let $\layerRV{\ell}$
be a 
random vector 
sampling which  yields a specific configuration $\bb{a}_\ell$ for layer $\ell$. The probability of the network configuration $\bb{a}$ is given by
\begin{equation}
    p(\bb{a} | \bb{\alpha}) = \prod_{\ell=1}^L \Pr( \layerRV{\ell} = \bb{a}_\ell ).
\end{equation}
Note that the latter probability depends on the $\hat{\bb{\alpha}}_\ell$'s, which, in turn, depend on the $\bb{\alpha}_\ell$'s.
The neural network is thus fully defined by the configuration $\networkConfig$ 
and the 
weights $\networkWeights$. 

Let $\lossConfig$ denote the loss of the particular configuration $\networkConfig$ with the weights $\networkWeights$. The expected loss over all network configurations with the same weights is given by
\begin{equation}
J(\bb{\alpha} ; \networkWeights) = \mathbb{E}_{\hat{\bb{\alpha}}} \lossConfig = \sum_{\bb{a}} p(\bb{a} | \bb{\alpha}) \lossConfig,
\end{equation}
where the sum is taken over all possible configurations.
The goal of the search is to minimize the latter loss over $\bb{\alpha}$ and $\networkWeights$, which is carried out using gradient steps. 

Note that compared to the regular neural network training (optimization over $\networkWeights$ with a single configuration), the number of additional optimization variables ($\bb{\alpha}$) remains relatively modest. 
In sharp contrast, the number of 
configurations required to compute 
$J$ is exponentially large, as shown in Lemmas \ref{multi_loss_pdv} and \ref{bi_loss_pdv} in the Appendix. For this reason, we approximate the gradient $\bb{g} = \nabla_{\bb{\alpha}} J$ by sampling a 
subset $S$ of possible configurations:
\begin{equation}  \label{eq:est}
    \opDerivative{i} \approx \opDerivativeEstimator{i} = \frac{1}{|S|} \sum_{\networkConfig \in S} \lossConfig \cdot (\opSample{i} - \nFiltersLayer \cdot \opProb{i}). 
\end{equation}
The 
sample size $\abs{S}$ governs the tradeoff between the 
complexity of the training and the estimator variance. 
The gradient with respect to the weights $\networkWeights$ is computed as usual.

\paragraph{Loss} 
We use bit operations (BOPs) \cite{baskin2018uniq} as a complexity metric in the case of quantization. BOPs refers to the number of bit operations needed to perform inference. Since the bitwidth of operands might be different, we extend the definition of \citet{baskin2018uniq} to this case.  The exact derivation of the BOPs metric is provided \cref{app:bops}. Neither FLOPs nor BOPs 
predict the runtime of the network \cite{tan2018mnasnet,cai2018proxylessnas,li2019partial}, but can still be used as a proxy to 
the performance.

Let us denote the complexity of layer $\ell$  by $\bops_\ell \left(\vidx\right)$; the metric is defined in \cref{eq:complexity} in the Appendix for the quantized CNN case, and simply equals to the MAC count of the layer in all other cases. We define the computational complexity loss as
\begin{equation}
\begin{split}
\bopsLoss %
\triangleq %
\sigma \left({
\sum_{\ell=1}^L
{
    \bops_\ell(\vidx)
}
}\right),
\end{split}
\end{equation} 
where $\sigma$ is some 
increasing function. 
Note that $\bopsLoss$ only depends on the network configuration 
and penalizes configurations $\networkConfig$ with high complexity. In particular, $\sigma$ can be a function of the ratio between $\networkConfig$ arithmetic complexity and the complexity of some target homogeneous configuration, which allows to set a target complexity for the search.

The combined loss $\lossConfig$ appearing in (\ref{eq:est}) is a linear combination of the standard loss used to train the network w.r.t. the weights, $\ceLoss$, 
and the complexity loss,
\begin{equation}
    \lossConfig = \ceLoss + \lambda \cdot \complexityLoss,
\end{equation}
embodying the tradeoff between the network accuracy and complexity.

\section{Quantized NAS}
\label{sec:quant_exp}
Quantization was one of the evaluated compression methods.
We used ResNet-20 \cite{he2016resnet} on CIFAR-10 
\cite{krizhevsky2009cifar}. The network was quantized with NICE \cite{baskin2018nice}, with the operations set $\layerOpsSet$ consisting of tuples $\bitwidthTuple$ of weight and output activation bitwidths, respectively.
$\layerRV{\ell} \sim \multinomial \left( \nFiltersLayer, \left( \opProb{1}, \opProb{2}, \dots, \opProb{|\layerOpsSet|} \right) \right)$ is multinomial random variable, with the probabilities $\bb{\opProb{}}$ obtained from 
$\bb{\alpha}$  
using softmax.

Sampling a layer configuration $\layerConfigDetailed$ 
induces a specific structure over the 
filters, shown in \cref{fig:multinomial_filter_wise_layer} in the Appendix. For $\layerOpsSet = \{ t_1, t_2, \dots, t_{|\layerOpsSet|} \}$, 
we apply quantization with bitwidth tuple $t_1$ on the first $\opSample{1}$ filters, $t_2$ on the next $\opSample{2}$ filters and so on.



In our experiments, we selected the 
set $\layerOpsSet = \qty{ (2,2), (2,4), (3,3), (8,8) }$ for all the layers. A few configurations were trained multiple times under the same conditions. 
We conclude that though the search yields 
well-performing configurations (\cref{fig:quant_results,app:quantres}), 
the variance of the accuracy 
is high, as shown in \cref{fig:NICE_variance}. 
Thus, it is impossible to establish 
whether the configurations would be good in a different realization.


\section{Slimmable NAS}
\label{sec:pruning_exp}

Another compression method we considered is a reduction of number of filters in convolutional layers. In particular, we used slimmable networks framework  \cite{yu2018slimmable, yu2019universally}, in which networks with the same architecture but different amount of filters are trained simultaneously with the same weights. The operations set $\layerOpsSet = \qty{ 1, \dots , \nFiltersLayer }$ represents the number of filters in a layer $\ell$. We set $\layerRV{\ell} \sim \binomial \left( \nFiltersLayer-1 , \opProbSingle \right)$ as binomial a random variable, and use a sigmoid normalization of the distribution parameters,
\begin{equation}
    \opProb{} = \binomialProb
\end{equation}
Sampling a configuration from $\layerRV{\ell}$ determines the number of filters in the layer.

\begin{figure}[htbp]
 \centering
     \begin{subfigure}[b]{0.48\linewidth}
        \includegraphics[width=\linewidth]{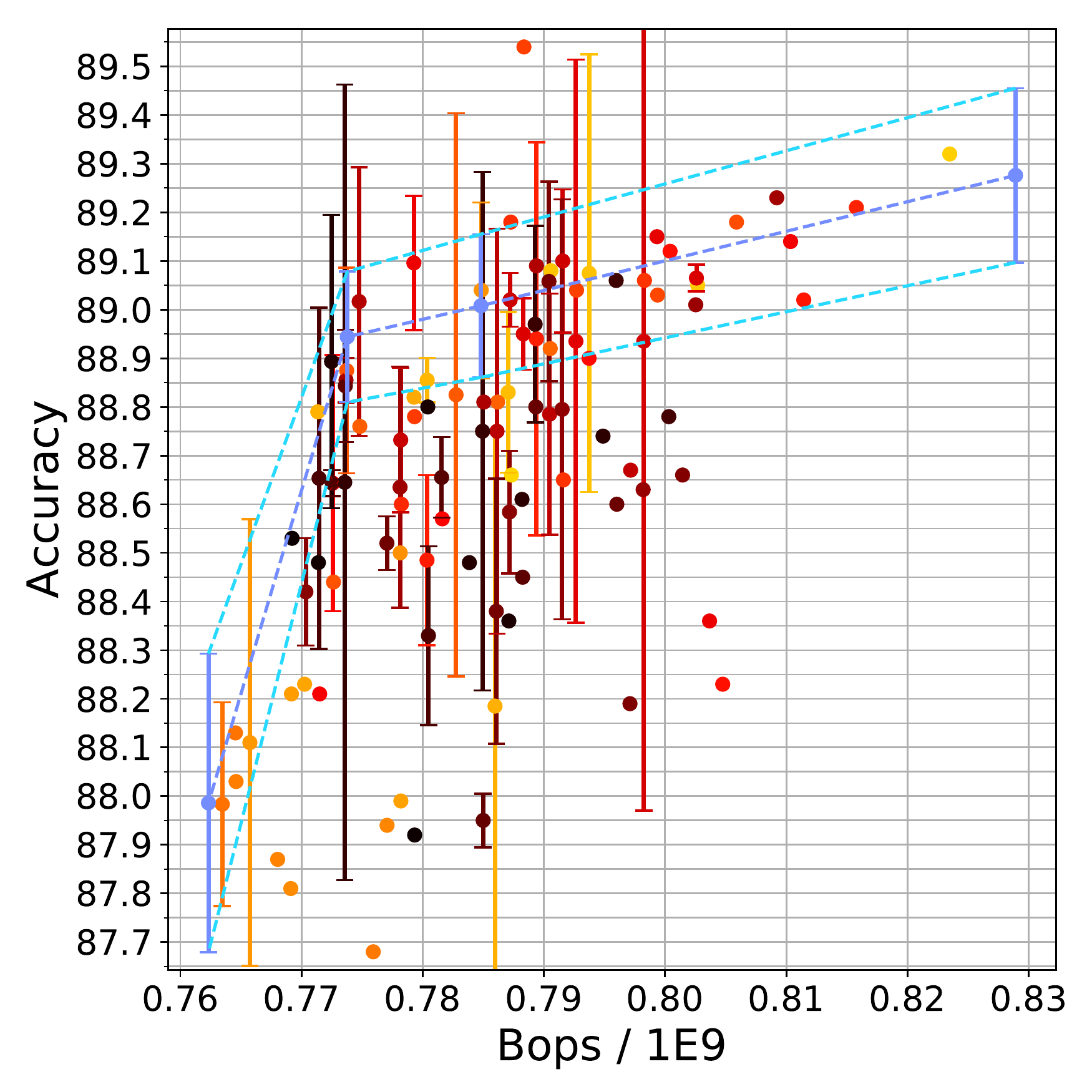}
        \subcaption{Results of grid search for quantization. 
        }
        \label{fig:NICE_variance}
    \end{subfigure}  \hfill 
     \begin{subfigure}[b]{0.48\linewidth}
        \includegraphics[width=\linewidth]{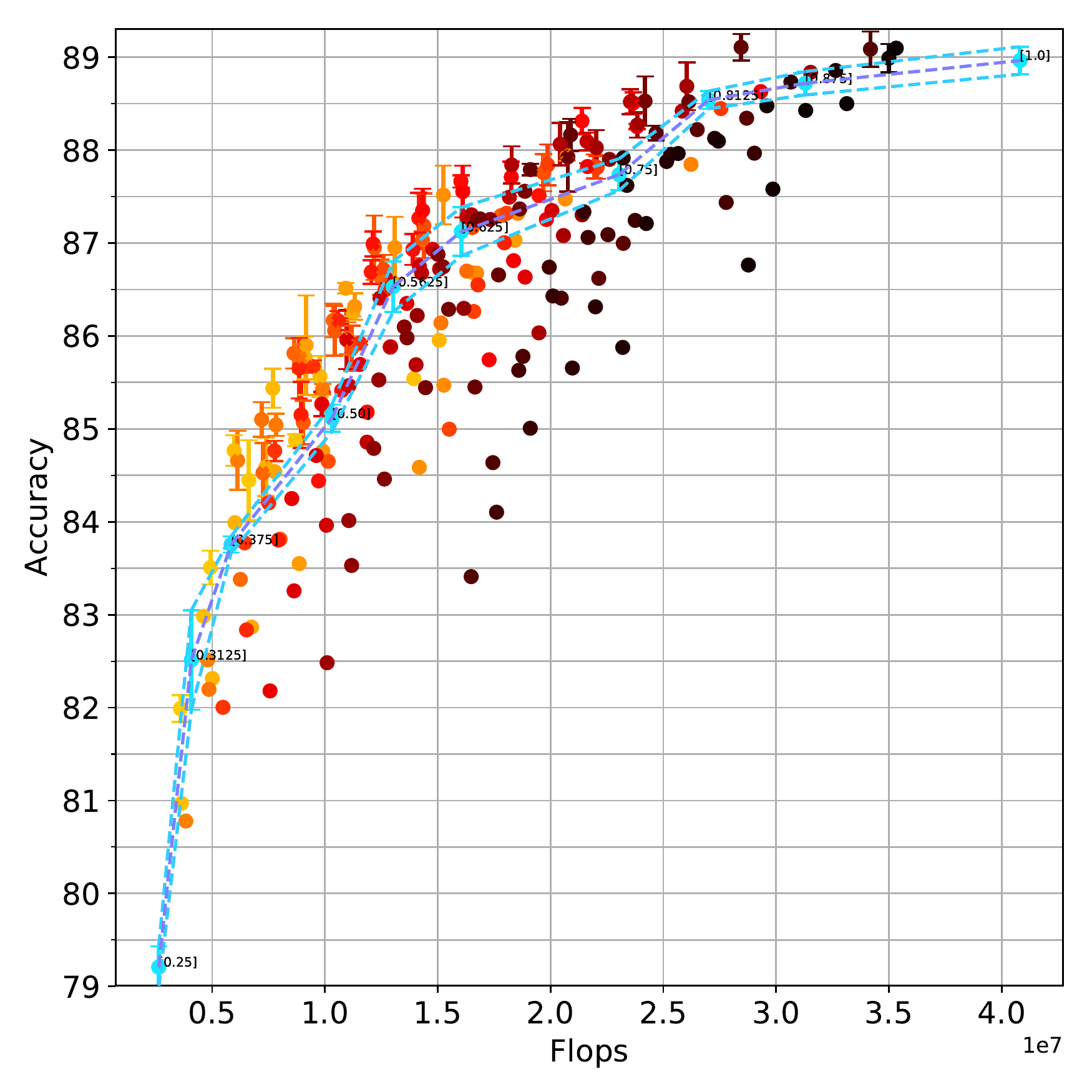}
        \subcaption{Results of grid search for pruning.}
        \label{fig:Slimmable_variance}
    \end{subfigure}
  \caption{Results of grid search in both cases. Blue line connects homogeneous configurations, colored points are heterogeneous configurations. Error bars are for $0.6827$ confidence interval.  More details in \cref{fig:NICE_variance_ext,fig:Slimmable_variance_ext}.}
  \label{fig:grid_search}
  \vspace{-0.2in}
\end{figure}

Similarly to \cref{sec:quant_exp}, we explored the search space by evaluating ResNet-20 configurations 
with setup described in \cref{app:searchspace}. As shown in  \cref{fig:Slimmable_variance}, the variance is relatively low, though it is still higher for heterogeneous configurations. 
Points with statistically significant improvement over homogeneous configuration were also found.

\paragraph{Basic search method}
\label{method_I}
At each iteration, we sample a set of configurations $\configSubset{k}$ from  current distribution $\prob^{k}$ 
for gradient estimation. To improve the loss evaluation 
we duplicate the current network weights and fine-tune each configuration $\networkConfig \in \configSubset{k}$ for 5 epochs.

We define the expected configuration $\expectedConfig{}$ such that $\expectedConfig{}_l = \round \qty( \mathbb{E} \qty[ \layerRV{\ell} ])$. The network weights are trained
over 5 configurations:  4 homogeneous ones ($\qty{ 0.25, 0.5, 0.75, 1.0 }$) and one defined by $\expectedConfig{}$.
Since samples from $\prob^{k}$ would be close to the expectation of their distribution,
$\networkWeights$ should be a better starting point to train the sampled configurations.

\paragraph{Resetting the $\bb{\omega}$}
\label{method_II}
We noticed that after few iterations of 
the network weights updates 
on $A \cup \expectedConfig{k}$, 
the validation loss of $\expectedConfig{k}$  was high
compared to the homogeneous configurations in $A$. 
We conjectured that the network overfits to the homogeneous configurations which are kept same while $\expectedConfig{k}$ changes. 
To avoid the overfitting, we reinitialize $\networkWeights$ after each iteration.
Additional changes, detailed in \cref{app:searchii}, were done.

\paragraph{Disabling weight-sharing}
\label{method_III}
In addition, the overly short fine-tuning 
leads to inaccurate configuration evaluation.
Thus  instead of sharing and fine-tuning $\networkWeights$  we trained each configuration $\networkConfig$, individually, with individual weights set $\configWeights$, from scratch.

\begin{figure}[htbp]
  \vspace{-0.17in}
 \centering
     \begin{subfigure}[b]{0.48\linewidth}
        \includegraphics[width=\linewidth]{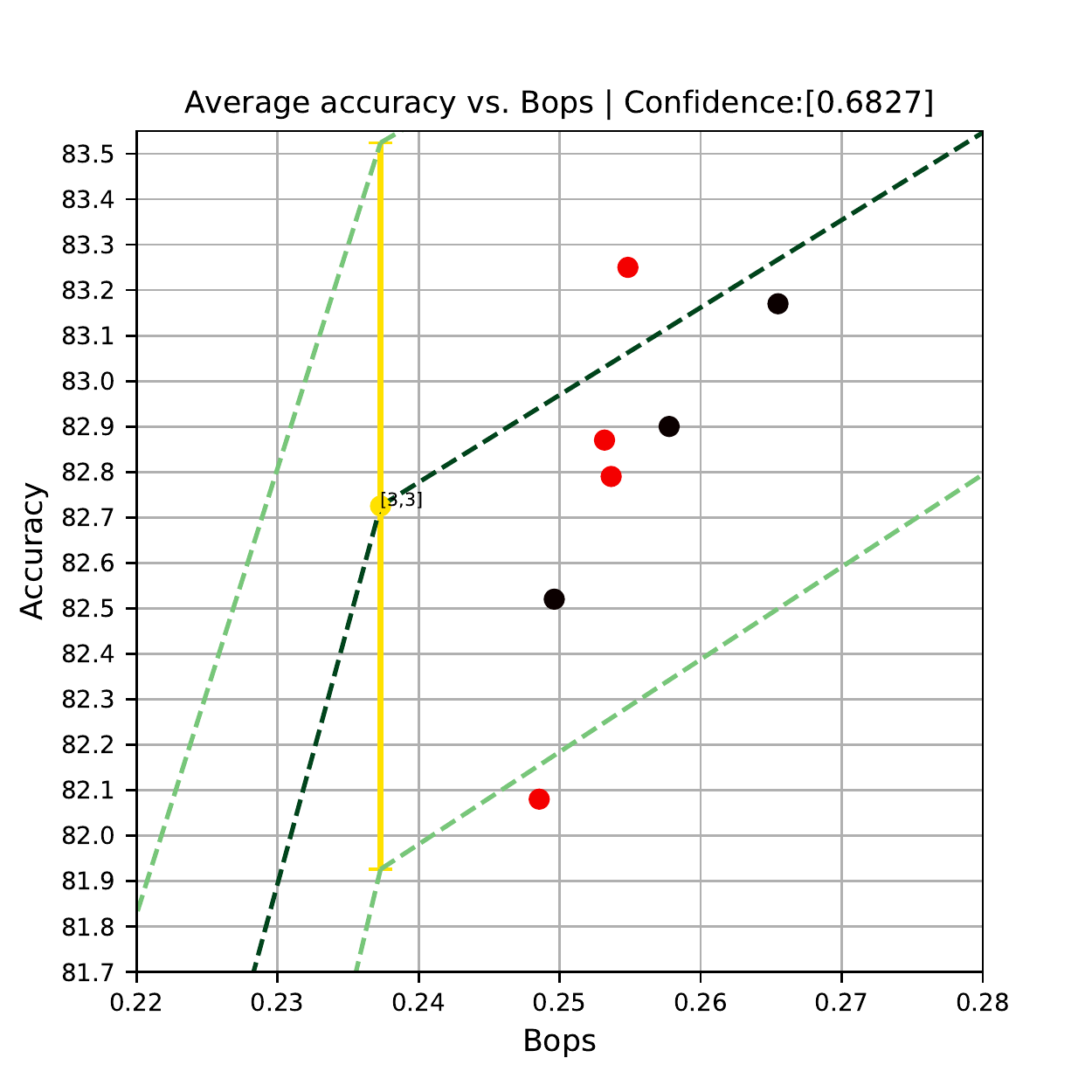}
        \subcaption{Results of search for quantization. 
        }
        \label{fig:quant_results}
  \vspace{-0.1in}
    \end{subfigure}  \hfill 
     \begin{subfigure}[b]{0.48\linewidth}
        \includegraphics[width=\linewidth]{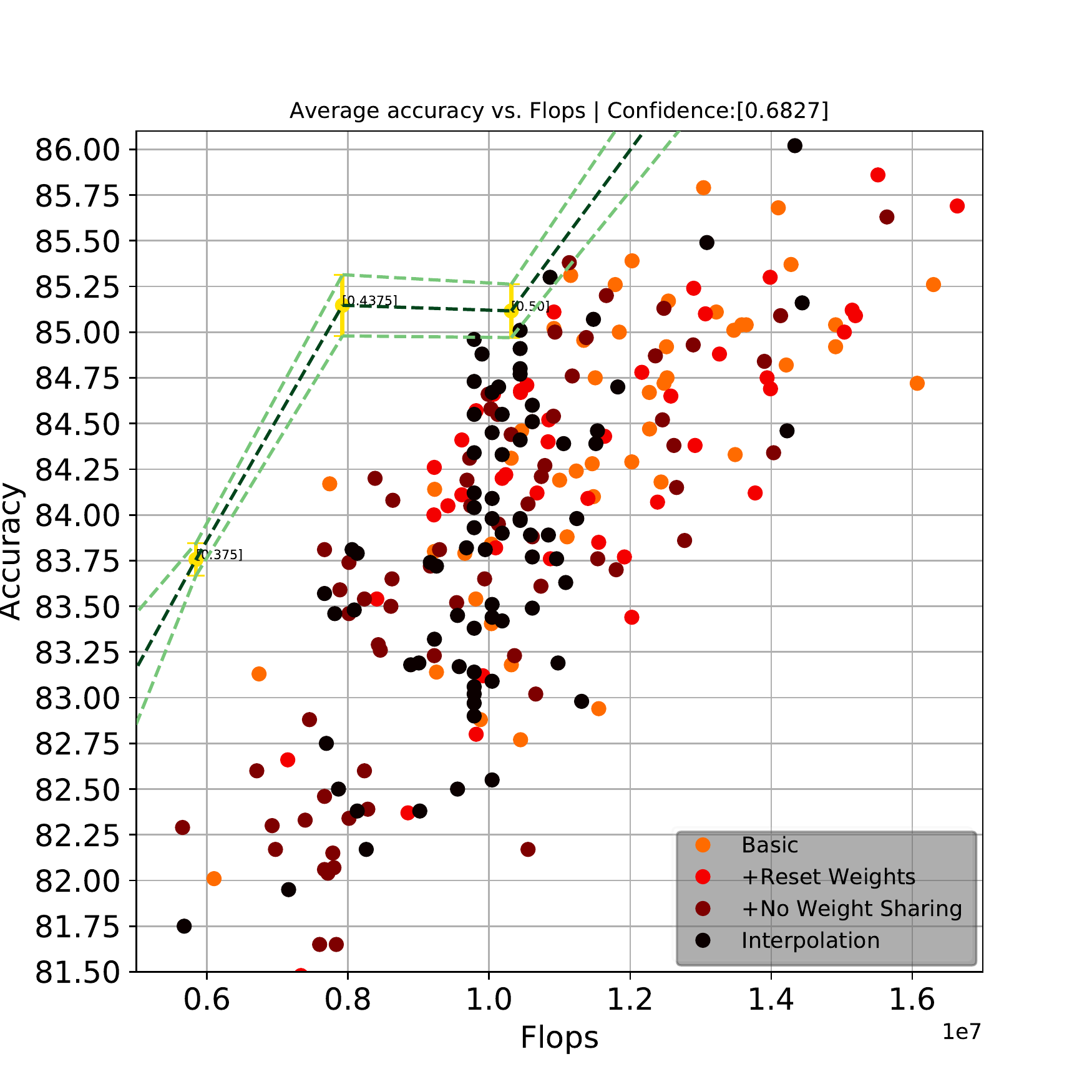}
        \subcaption{Results of search for pruning.}
        \label{fig:prun_results}
  \vspace{-0.1in}
    \end{subfigure}
  \caption{Results of search in both cases. Each point represents a configuration proposed by search. 
  More details in \cref{app:res}.}
  \label{fig:results}
  \vspace{-0.26in}
\end{figure}

\paragraph{Interpolation loss}
\label{method_IV}
To achieve the goal of improvement over homogeneous configurations, we tried to compare the heterogeneous configuration cross-entropy with the expected one,
by defining the loss as a difference from interpolation of known homogeneous configurations (details in \cref{app:searchiv}). The results are shown on \cref{fig:prun_results} and 
in \cref{app:prunres}.

\section{Conclusion}
\label{sec:conclusions}
In this paper, we studied the feasibility of using NAS-like algorithms, and in particular differentiable NAS \cite{liu2018darts}, for the reduction of CNN complexity by means of filter-wise quantization and layer-wise pruning. 
In both cases, we applied our method for ResNet-20 on CIFAR-10, on which it took only $36$ GPU-hours to converge. 

For filter-wise quantization, after acquiring nominal improvement over the homogeneous baseline, we found out that the variance of heterogeneous configurations is too high to warrant a significant improvement, which we confirmed using partial grid search. Unfortunately, previous studies on bitwidth allocation or 
architecture-quantization search \cite{wu2018mixed,elthakeb2018releq,chen2018joint,lou2019autoqb,guo2019single} did not report the variance of the results, making meaningful comparison impossible.
For layer-wise pruning, we obtained more stable results, with the grid search confirming the possibility of improvement over the baseline homogeneous configurations. However, the  heterogeneous configurations found by NAS did not significantly outperform the 
baseline.

We conclude that future work should 
focus on 
loss design
and
better loss estimators, 
such as Gumbel softmax \cite{jang2016categorical}. Successfully transferring the architecture to a more challenging use case 
(\eg 
ImageNet) 
remains another important challenge.

\acks{The research was funded by Hyundai Motor Company through HYUNDAI-TECHNION-KAIST Consortium, ERC StG RAPID and Hiroshi Fujiwara Technion Cyber Security Research Center. }

\vskip 0.2in
\clearpage

\bibliography{compress_nas}

\newpage
\appendix
\crefalias{section}{appsec}
\crefalias{subsection}{appsec}
\crefalias{subsubsection}{appsec}

\section{Configurations}

\begin{figure}[htb]
  \centering
    \includegraphics[width=0.25\linewidth]{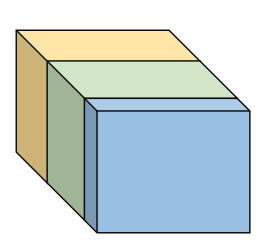}
  \caption{Layer structure induced by sampling a configuration from a multinomial random variable. Assume $\layerOpsSet = \{ t_1, t_2, t_3 \}$, we apply quantization with bitwidth $t_1$ on the blue filters, $t_2$ on the green filters and $t_3$ on the yellow filters.}
\label{fig:multinomial_filter_wise_layer}
\end{figure}

\subsection{Configuration of search space} \label{app:searchspace}
Similarly to \citet{ying2019bench}, we performed a grid search on a simplified search space. ResNet-20 \cite{he2016resnet} was chosen as a basic architecture. This architecture has three blocks of convolutional layers, with increasing number of filters and decreasing dimensions of features. To reduce the required resources, the number of filters was reduced to 16, 32, and 64 in each group. For the quantization search space we sampled few layer configurations. Each sampled layer configuration, denote as $\layerConfig{}$, induces a network configuration by setting each of the network layers configuration to $\layerConfig{}$. We train each network configuration 3 times under the same conditions. The stopping criteria is 150 consecutive epochs without new optimal validation accuracy. For the pruning search space we sampled blocks configurations triplets. Each triplet induces a network configuration, where each layer in a specific block has the same layer configuration as any other layer in the specific block. As we did for the quantization search space, we train each network configuration 3 times under the same conditions. The stopping criteria is also the same.
 
\subsection{Transfer learning to ImageNet}
We took 3 configurations with significant accuracy difference between them on CIFAR-10 and trained them on ImageNet. We found out that the configurations ranking on CIFAR-10 is different than their ranking on ImageNet, \ie a configuration might be optimal on CIFAR-10 but average on ImageNet. 
\section{Additional results analysis}



\begin{landscape}
\begin{figure}[htb]
  \centering
    \includegraphics[width=\linewidth]{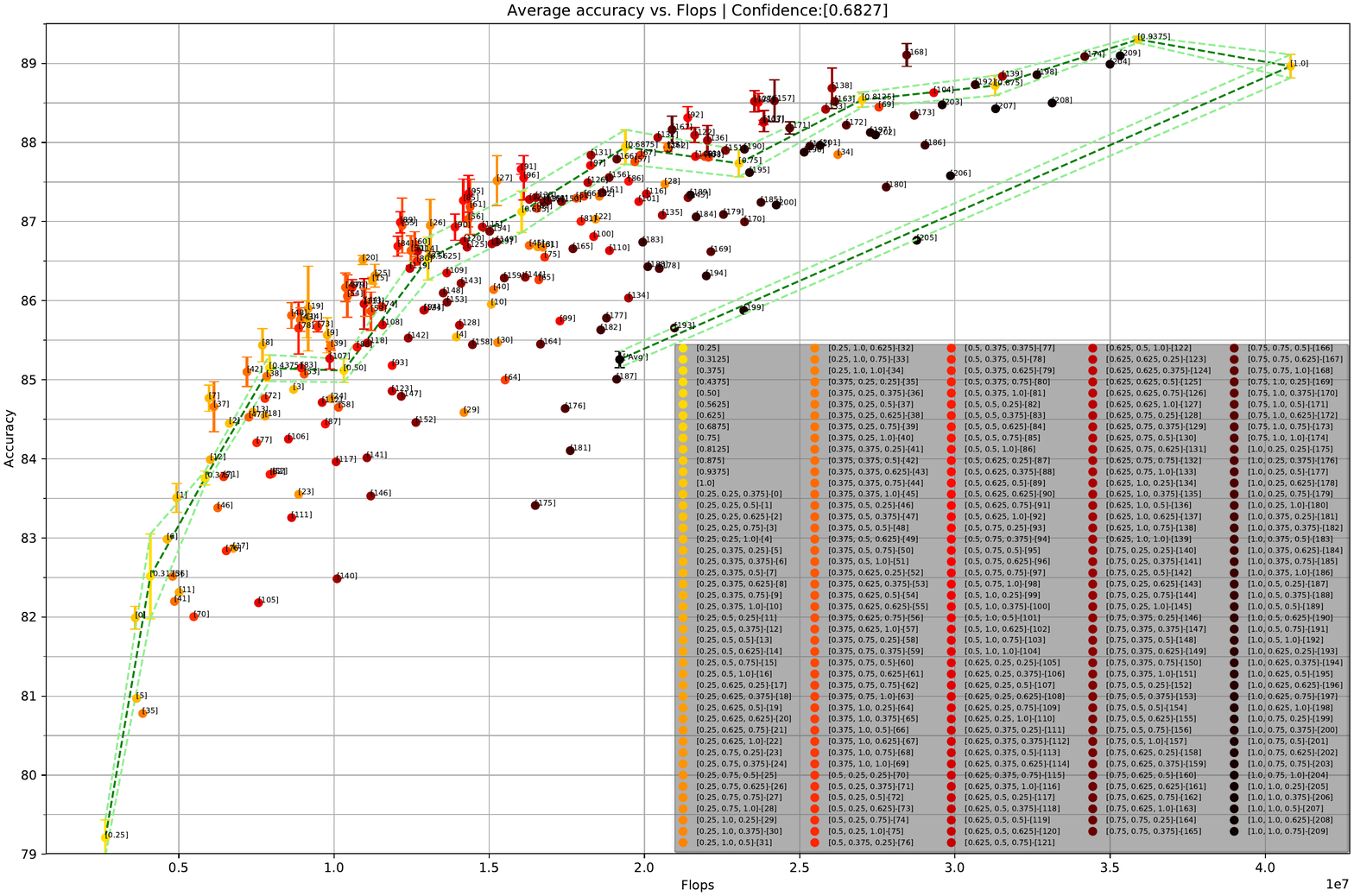}
    \caption{Slimmable validation accuracy variance exploration plot. Each point represents a different configuration. The points connected in dashed line are the homogeneous configurations. The error bar represents $0.6827$ confidence interval.}
\label{fig:Slimmable_variance_ext}
\end{figure}

\begin{figure}[htb]
  \centering
    \includegraphics[width=\linewidth]{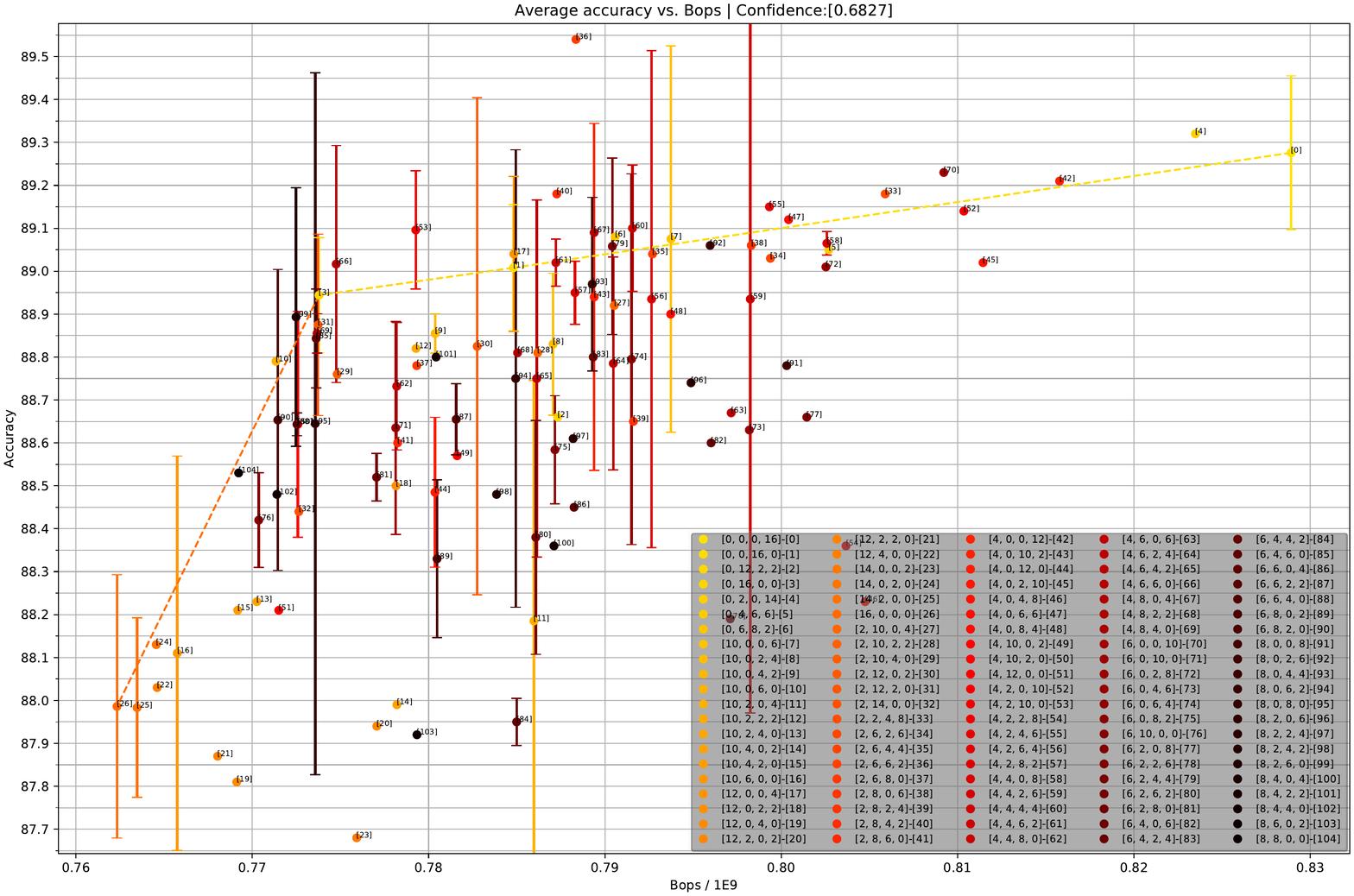}
    \caption{NICE validation accuracy variance investigation plot. Each point represents a different configuration. The points connected in dashed line are the homogeneous configurations. The error bar represents $0.6827$ confidence interval.}
\label{fig:NICE_variance_ext}
\end{figure}
\end{landscape}

\clearpage
\subsection{Results} \label{app:res}
\subsubsection{Quantization} \label{app:quantres}

\begin{figure}[htbp]
 \centering
     \begin{subfigure}[b]{0.48\linewidth}
        \includegraphics[width=\linewidth]{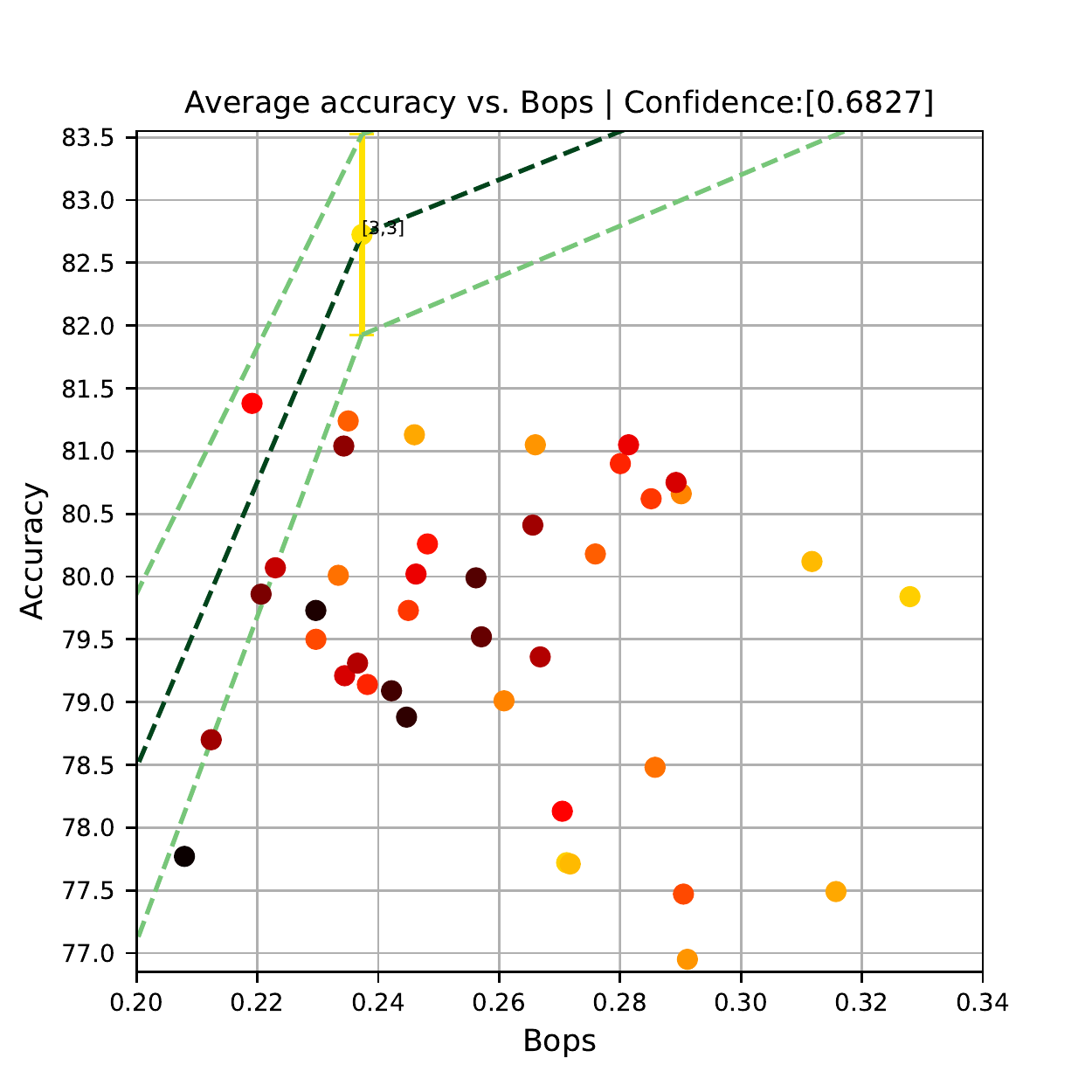}
        \subcaption{Homogeneous configuration target is (3,3).}
    \end{subfigure}  \hfill 
     \begin{subfigure}[b]{0.48\linewidth}
        \includegraphics[width=\linewidth]{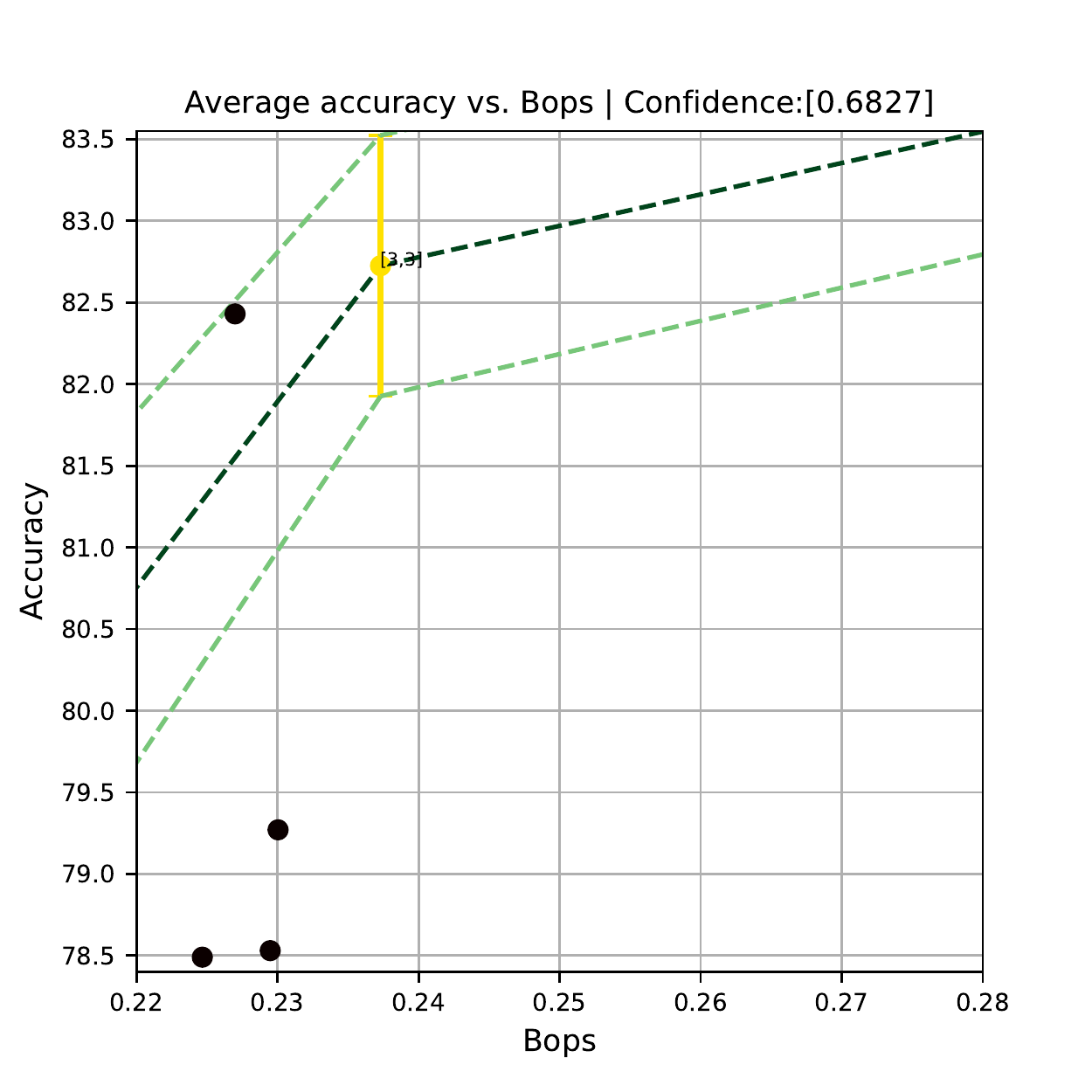}
        \subcaption{Homogeneous configuration target is (2,4).}
    \end{subfigure}\hfill 
  \caption{Additional results of search in quantization case. $\lambda=1$. $\layerOpsSet=\{(2,2),(2,4),(3,3),(8,8)\}$}
  \label{fig:q_results1}
\end{figure}
\begin{figure}[htbp]
 \centering
    \includegraphics[width=\linewidth]{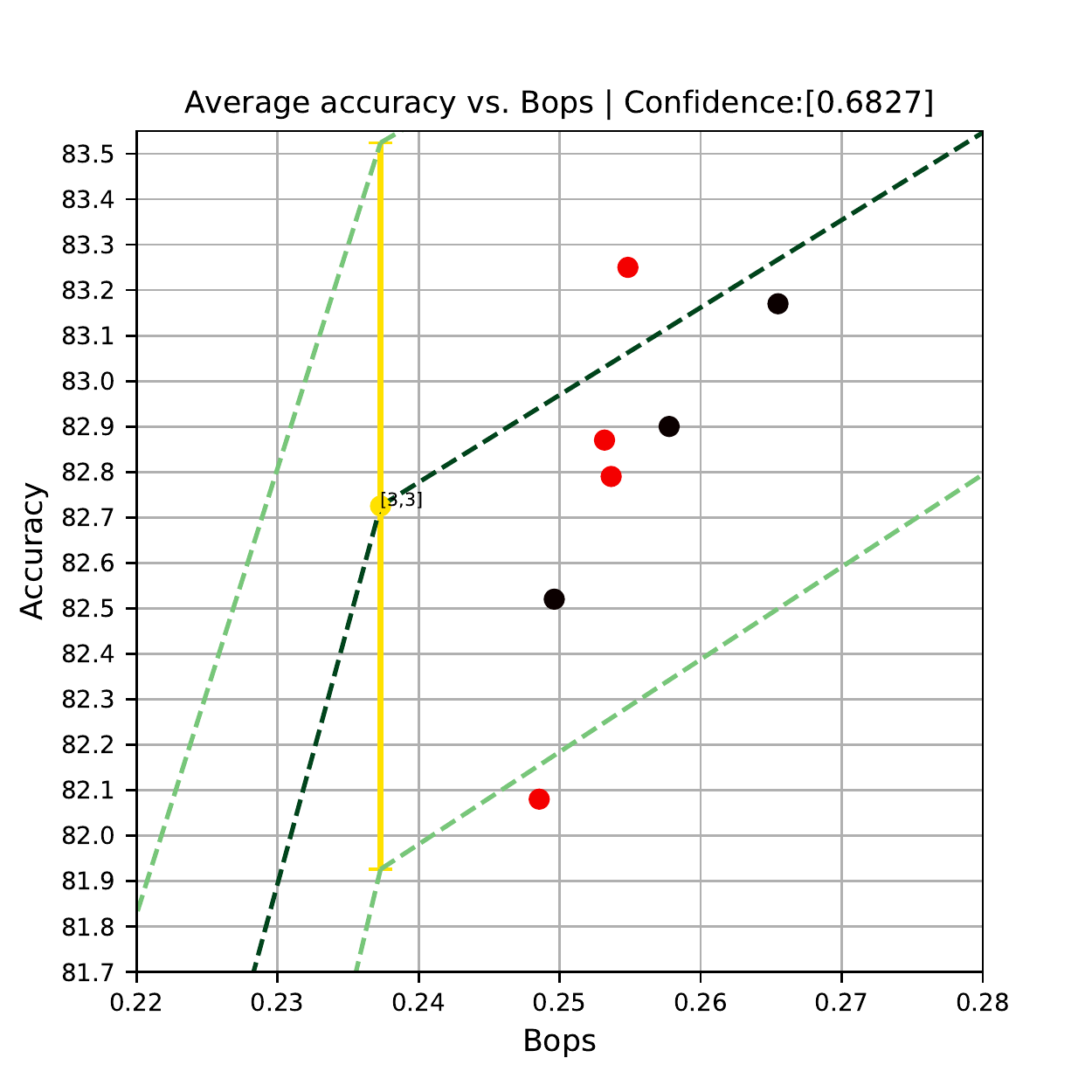}
  \caption{Additional results of search in quantization case. $\lambda=1$. Homogeneous configuration target is (3,3). $\layerOpsSet=\{(2,2),(3,3),(4,4),(8,3),(8,8)\}$.}
  \label{fig:q_results2}
\end{figure}

\begin{landscape}
\begin{figure}[htb]
  \centering
    \includegraphics[width=\linewidth]{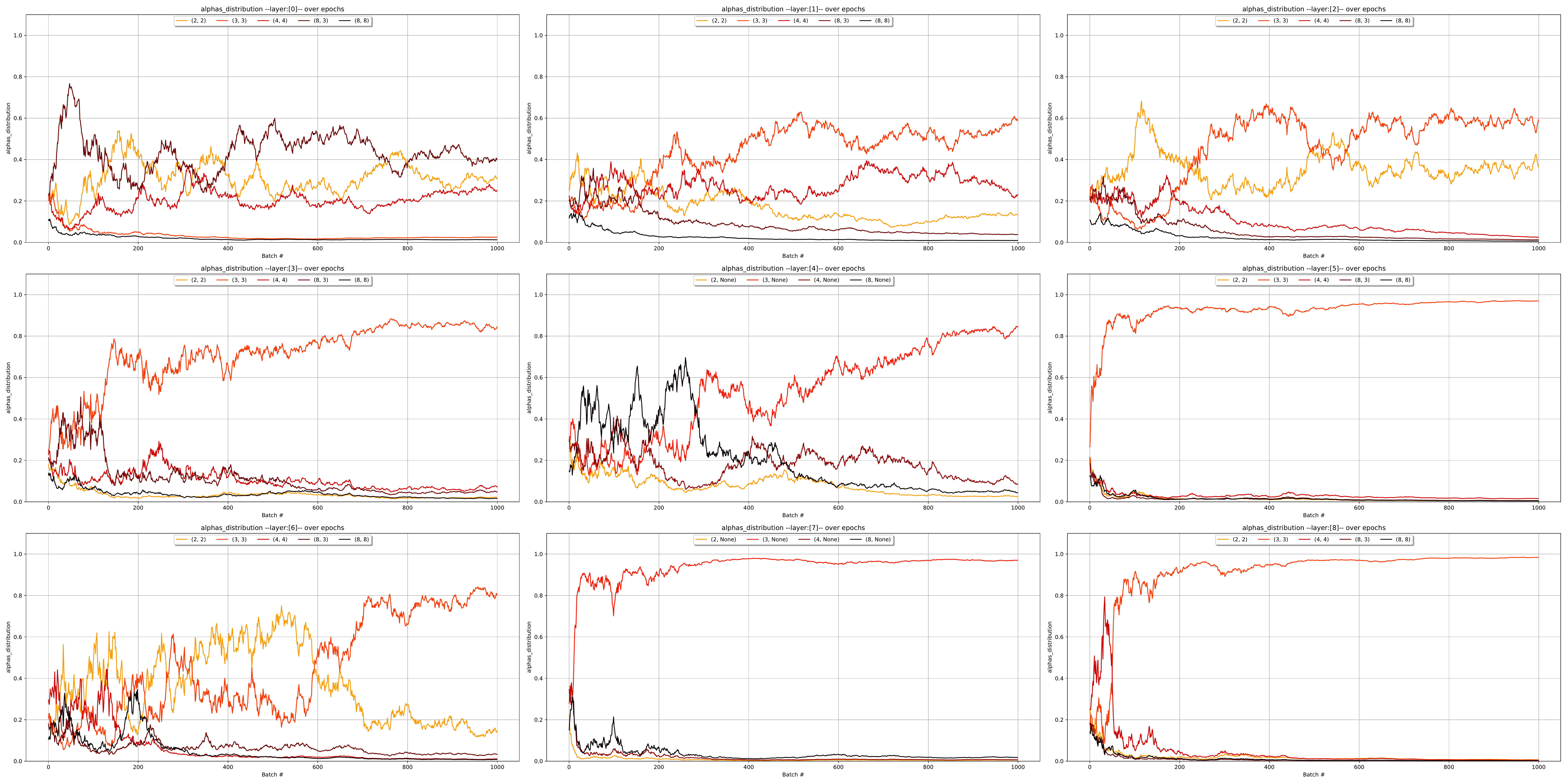}
    \caption{Convergence of $\bb{\alpha}$ as a function of time in quantization case}
\label{fig:quant_convergence}
\end{figure}
\end{landscape}

\clearpage
\subsubsection{Pruning}\label{app:prunres}

\begin{figure}[htb]
  \centering
    \includegraphics[width=\linewidth]{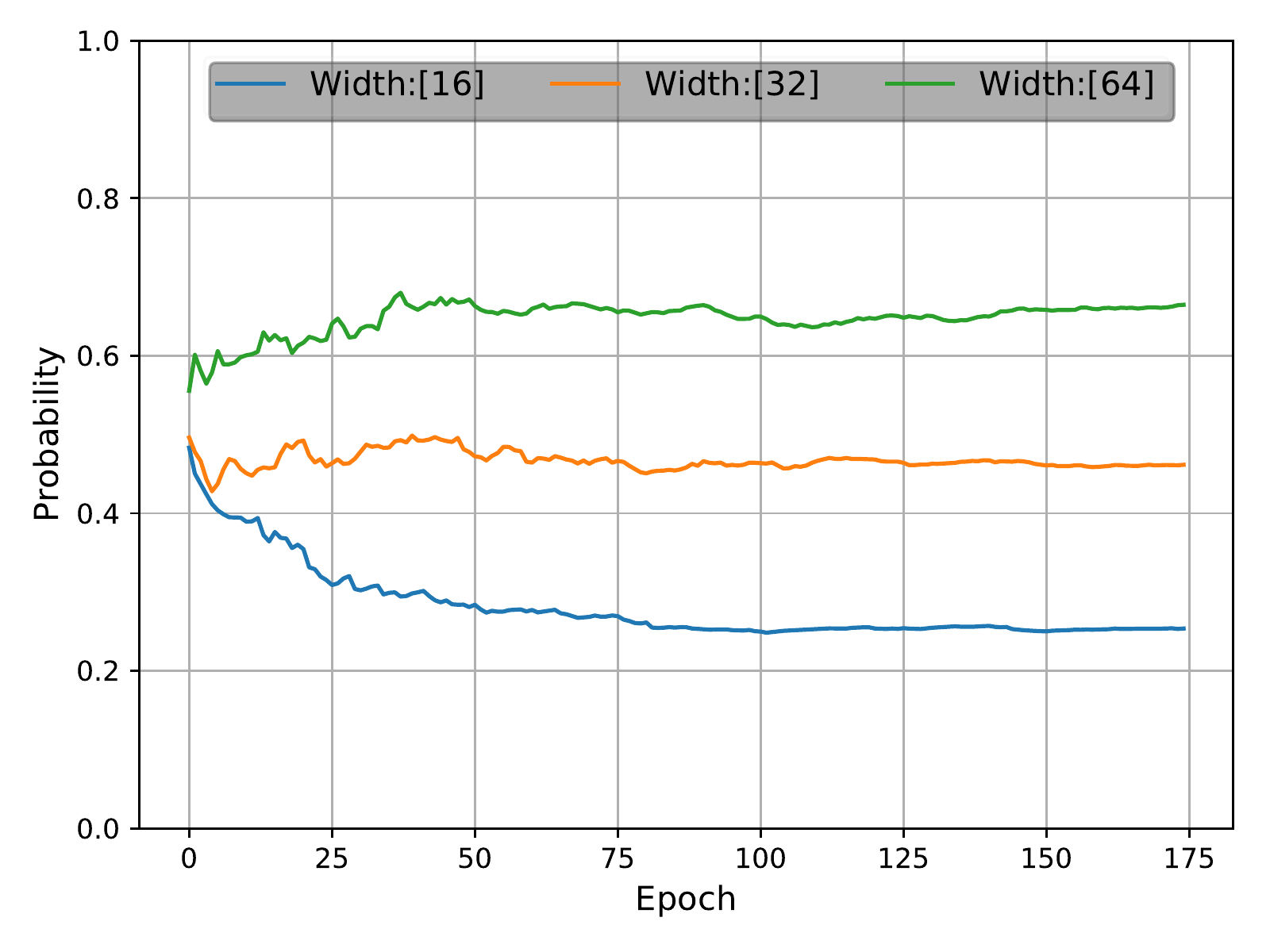}
    \caption{Convergence of $\bb{\alpha}$ as a function of time in pruning case}
\label{fig:prun_convergence}
\end{figure}

\begin{figure}[htbp]
 \centering
     \begin{subfigure}[b]{0.48\linewidth}
        \includegraphics[width=\linewidth]{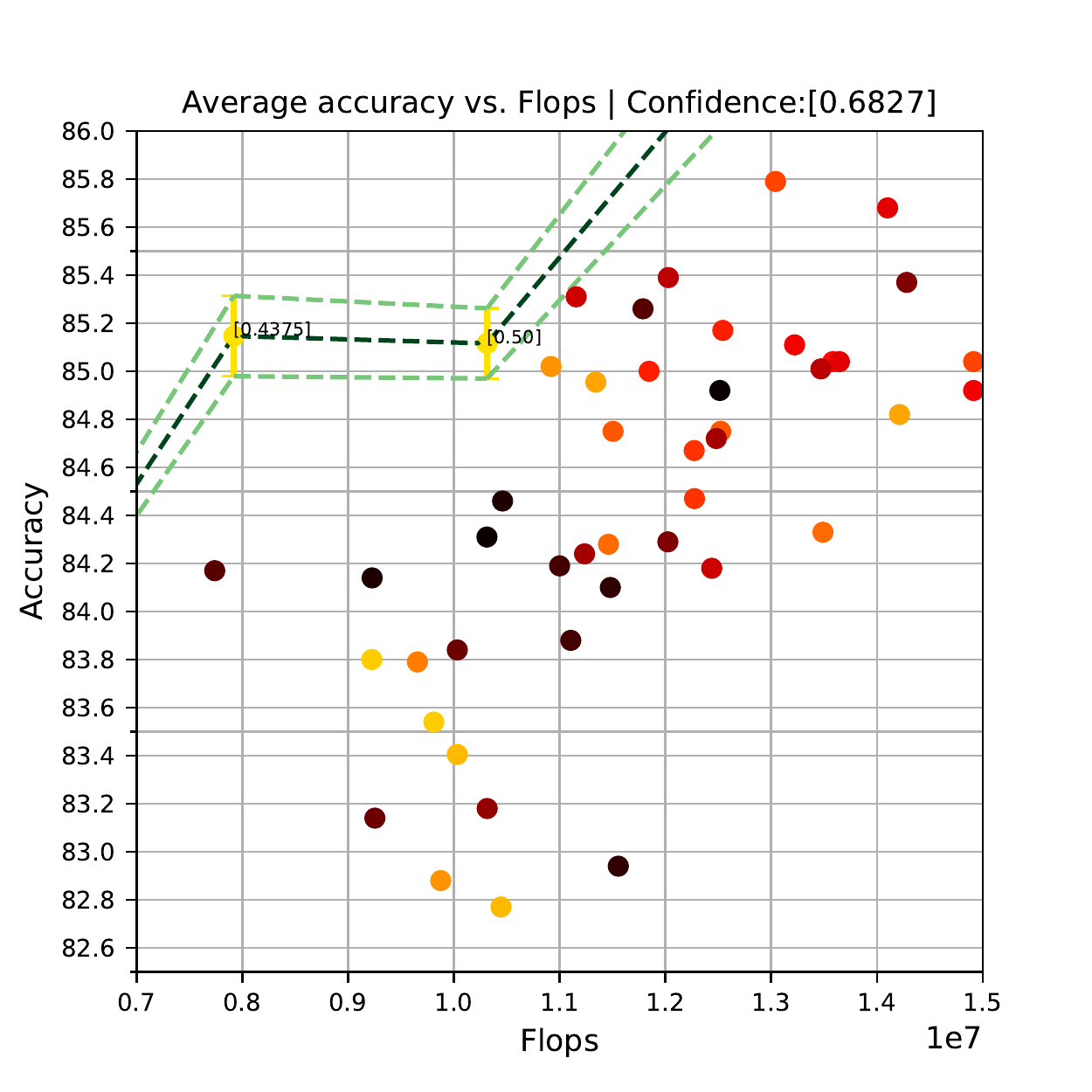}
        \subcaption{$\lambda=0.01,|S|=48$.}
    \end{subfigure}  \hfill 
     \begin{subfigure}[b]{0.48\linewidth}
        \includegraphics[width=\linewidth]{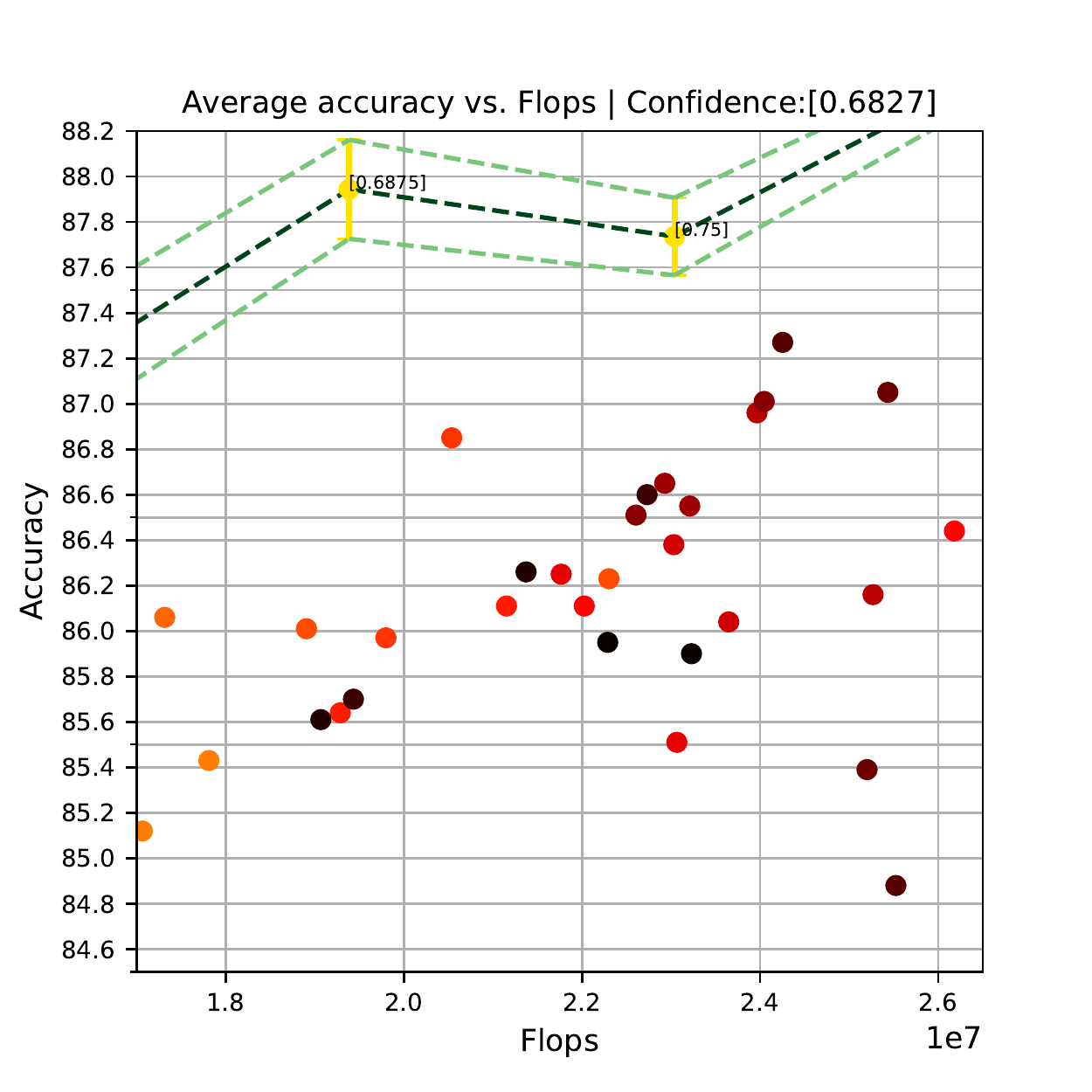}
        \subcaption{$\lambda=0.001,|S|=6$.}
    \end{subfigure}
  \caption{Results of basic search.}
  \label{fig:pm1_results}
\end{figure}

\begin{figure}[htbp]
 \centering
     \begin{subfigure}[b]{0.31\linewidth}
        \includegraphics[width=\linewidth]{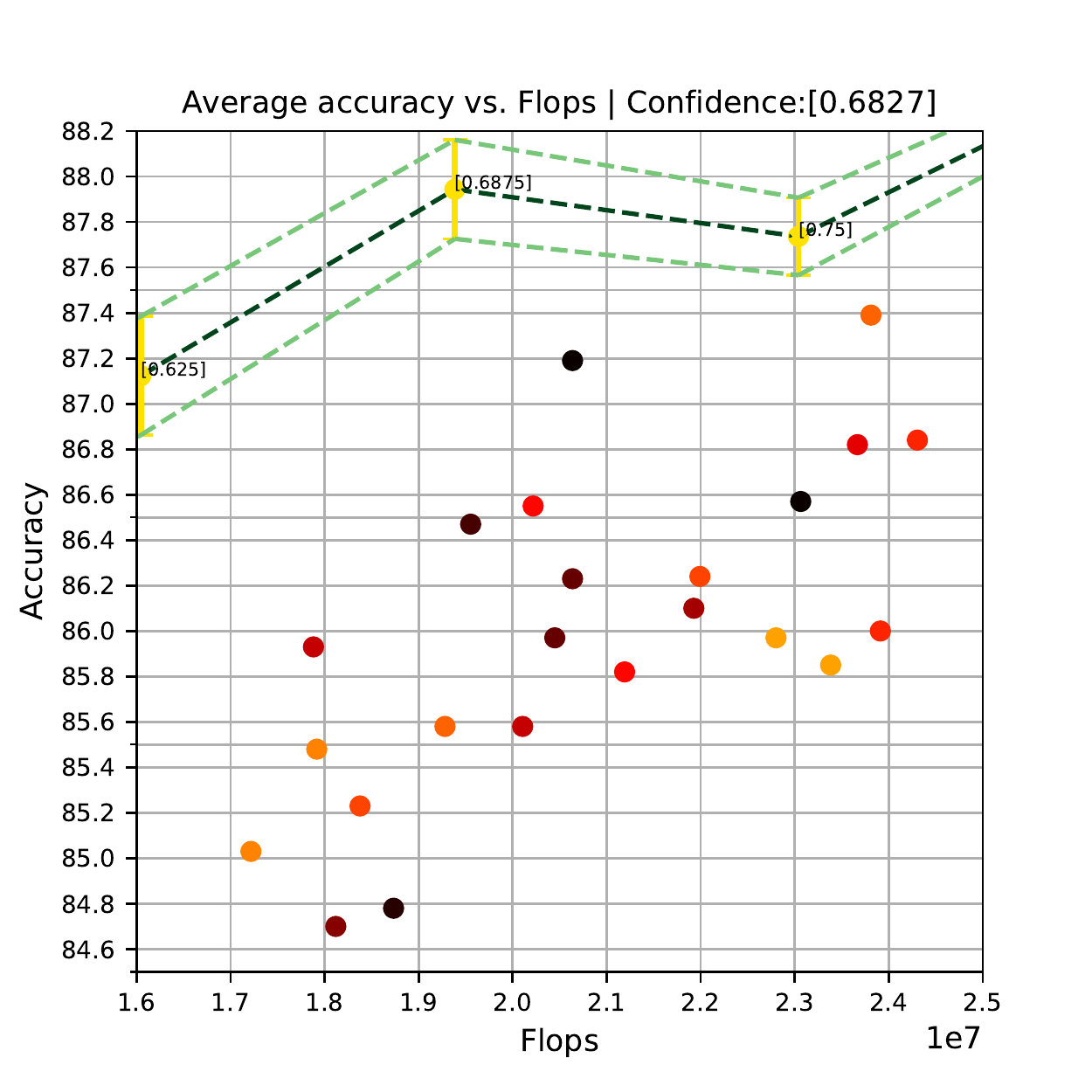}
        \subcaption{$\lambda=0.005,\trainWeightsInterval=20$.}
    \end{subfigure}  \hfill 
     \begin{subfigure}[b]{0.31\linewidth}
        \includegraphics[width=\linewidth]{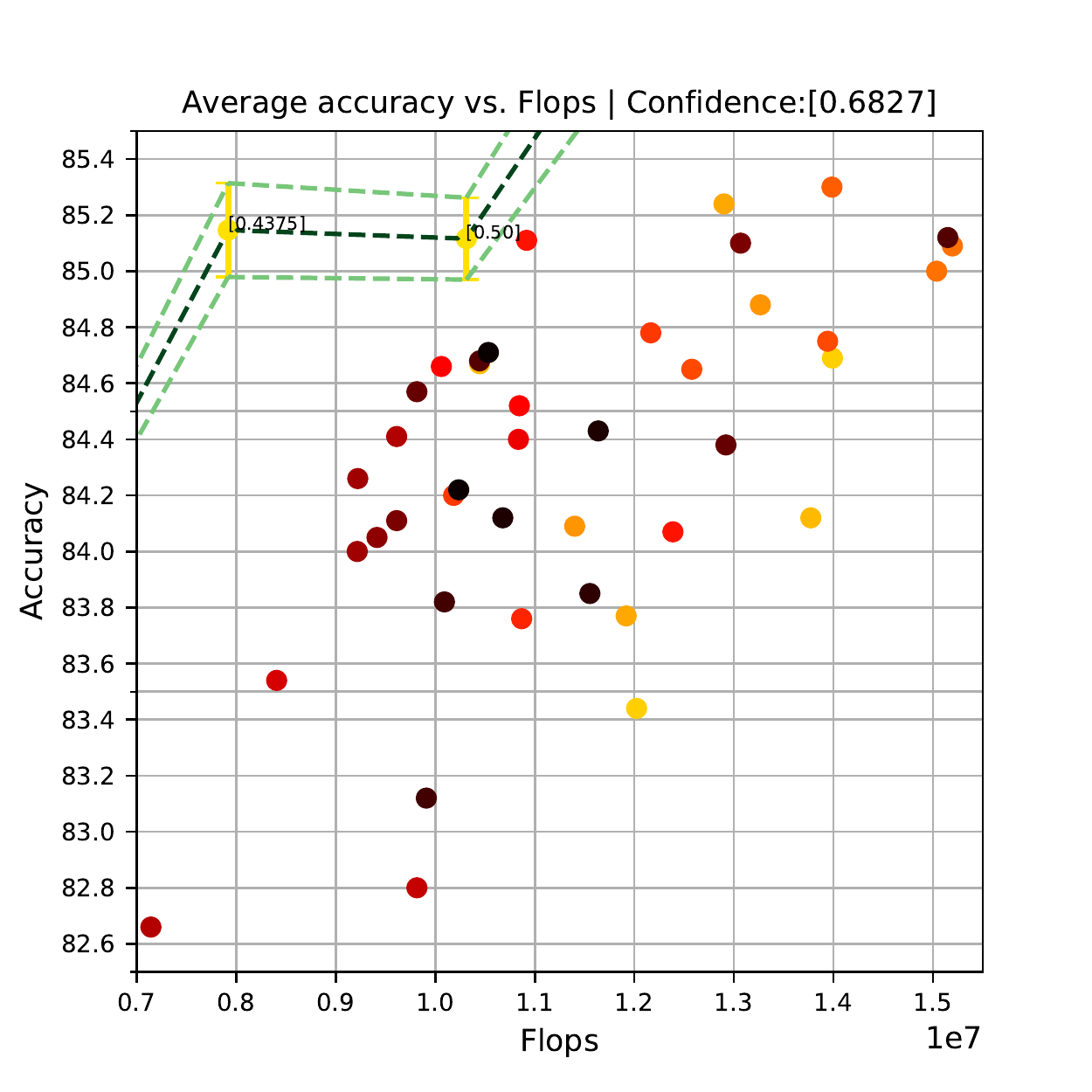}
        \subcaption{$\lambda=0.01,\trainWeightsInterval=10$.}
    \end{subfigure}\hfill 
     \begin{subfigure}[b]{0.31\linewidth}
        \includegraphics[width=\linewidth]{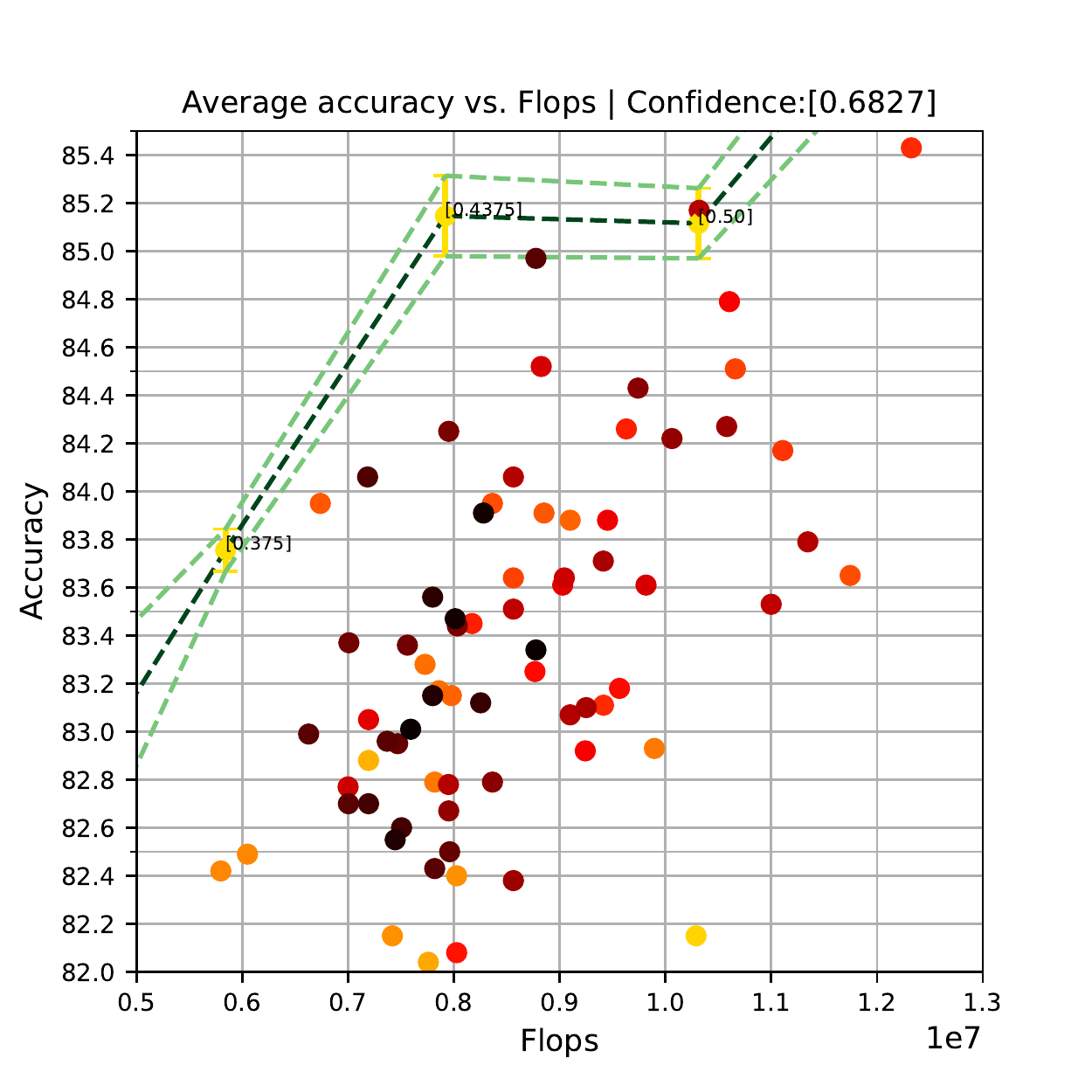}
        \subcaption{$\lambda=0.02,\trainWeightsInterval=10$.}
    \end{subfigure}
  \caption{Results of search with $\omega$ resetting. $|S|=6$.}
  \label{fig:pm2_results}
\end{figure}

\begin{figure}[htbp]
 \centering
     \begin{subfigure}[b]{0.48\linewidth}
        \includegraphics[width=\linewidth]{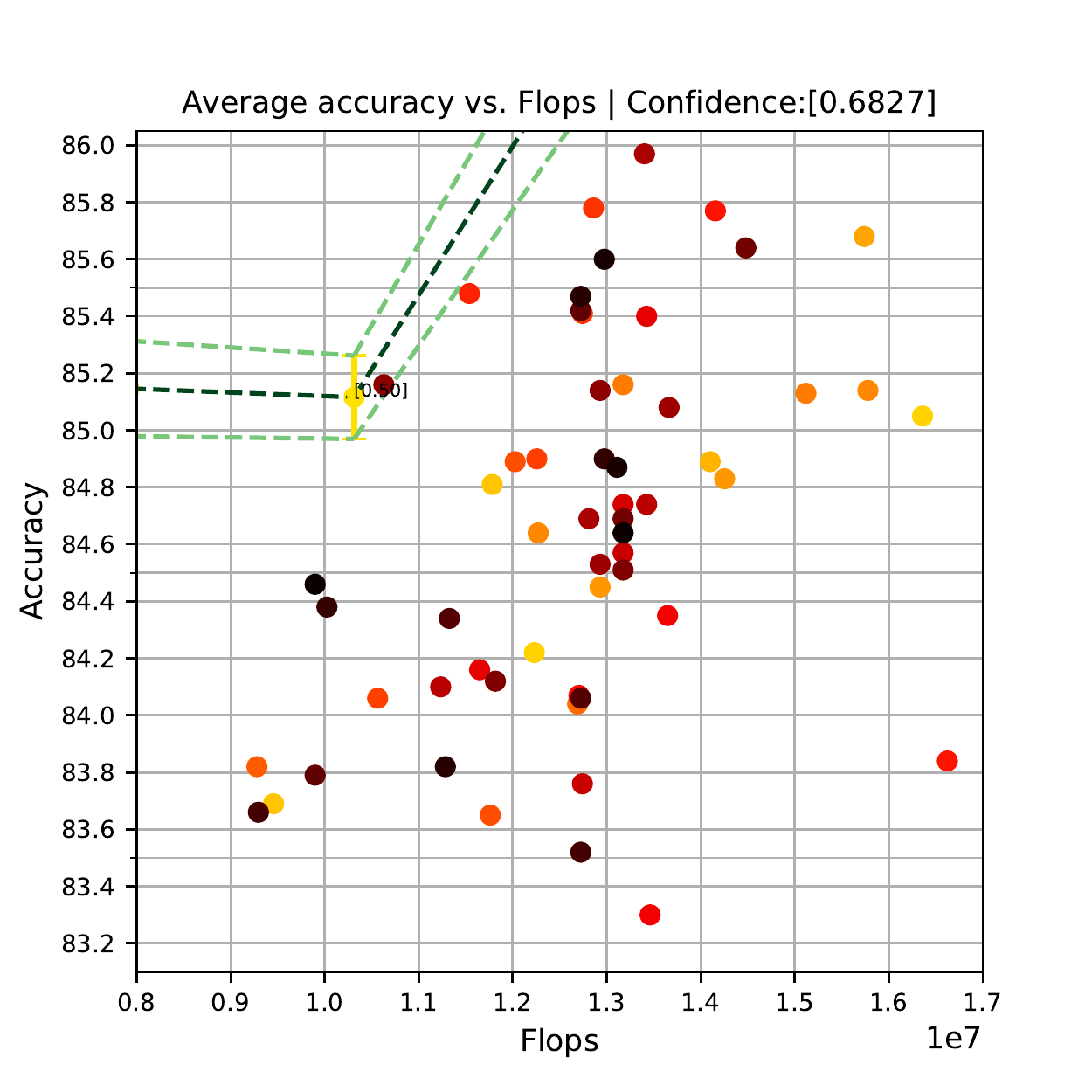}
        \subcaption{$\lambda=0.005,|S|=8$}
    \end{subfigure}  \hfill 
     \begin{subfigure}[b]{0.48\linewidth}
        \includegraphics[width=\linewidth]{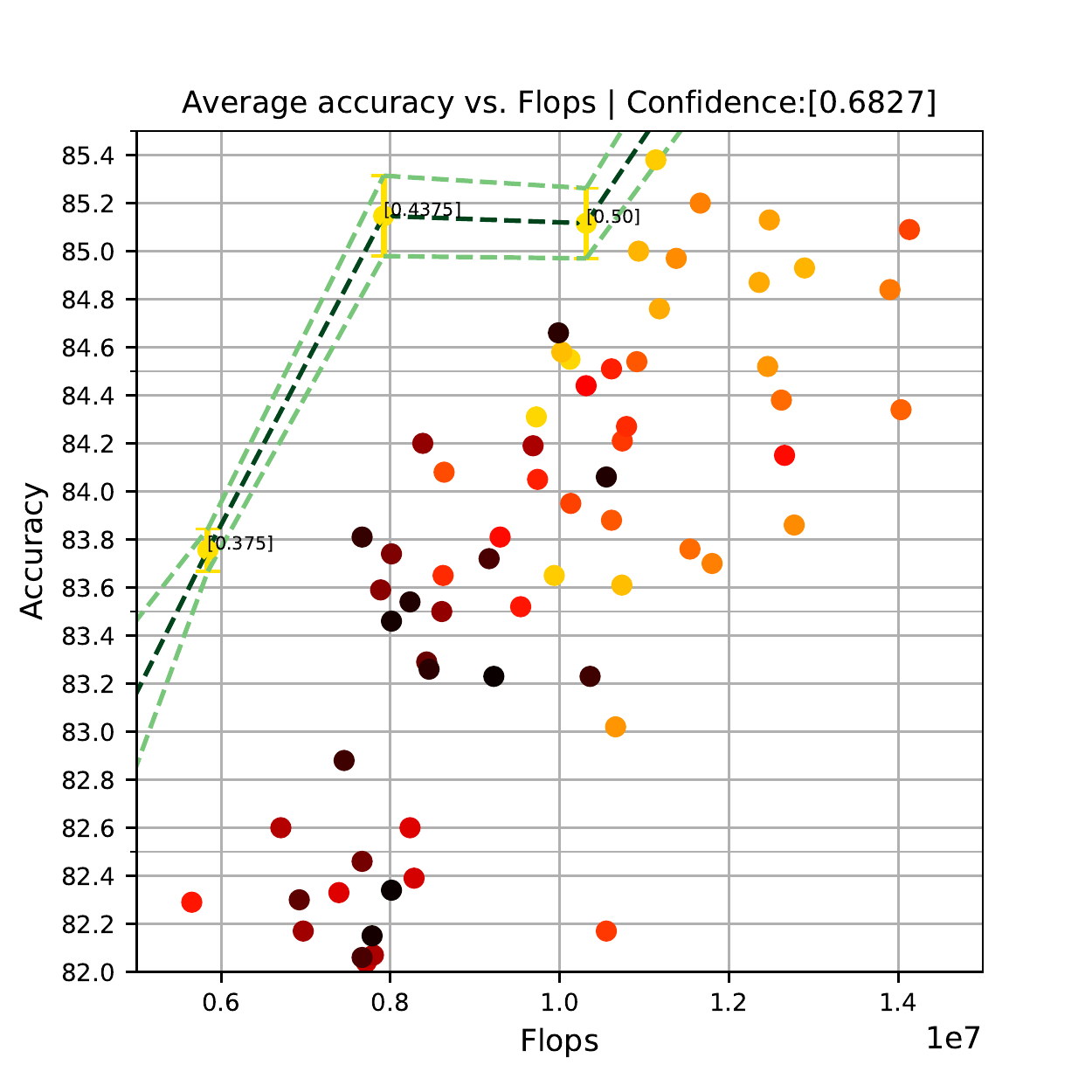}
        \subcaption{$\lambda=0.01,|S|=6$}
    \end{subfigure}
  \caption{Results of without weight sharing.}
  \label{fig:pm3_results}
\end{figure}

\begin{figure}[htbp]
 \centering
     \begin{subfigure}[b]{0.31\linewidth}
        \includegraphics[width=\linewidth]{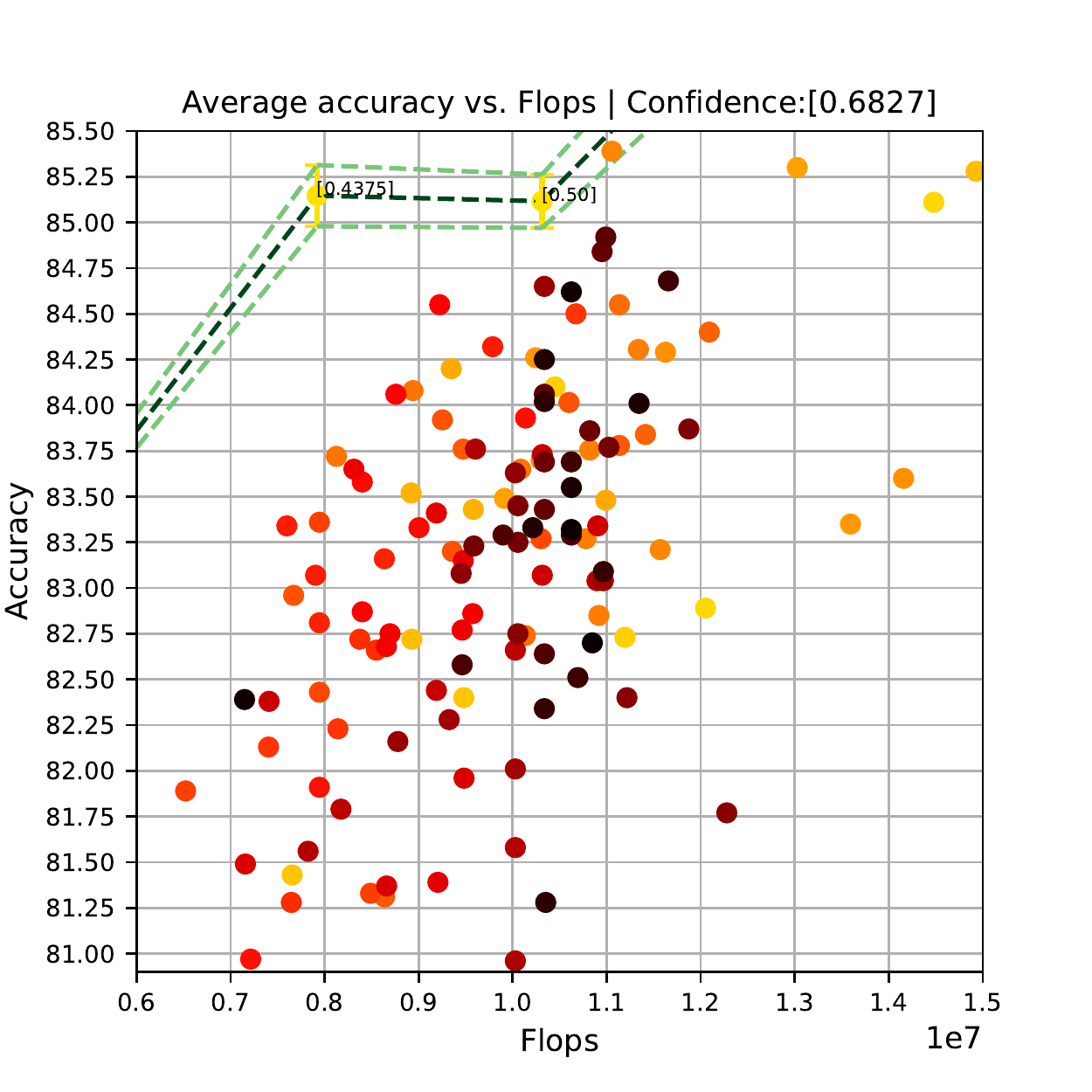}
        \subcaption{Learning rate = 5E-4. Initial distribution is 0.5, \ie $\opProbSingle=0.5$.}
    \end{subfigure}  \hfill 
     \begin{subfigure}[b]{0.31\linewidth}
        \includegraphics[width=\linewidth]{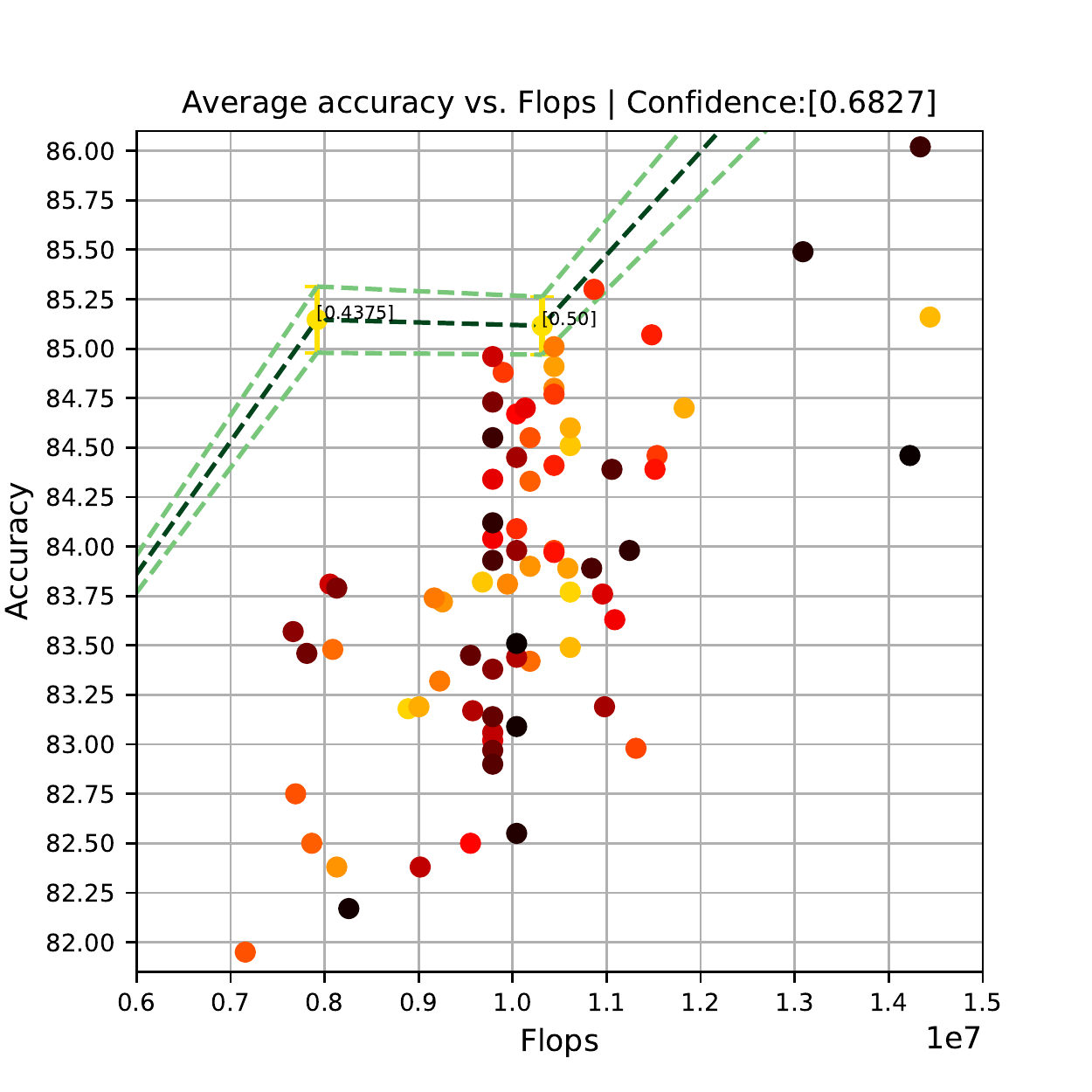}
        \subcaption{Learning rate = 1E-4. Initial distribution is 0.5, \ie $\opProbSingle=0.5$.}
    \end{subfigure}\hfill 
     \begin{subfigure}[b]{0.31\linewidth}
        \includegraphics[width=\linewidth]{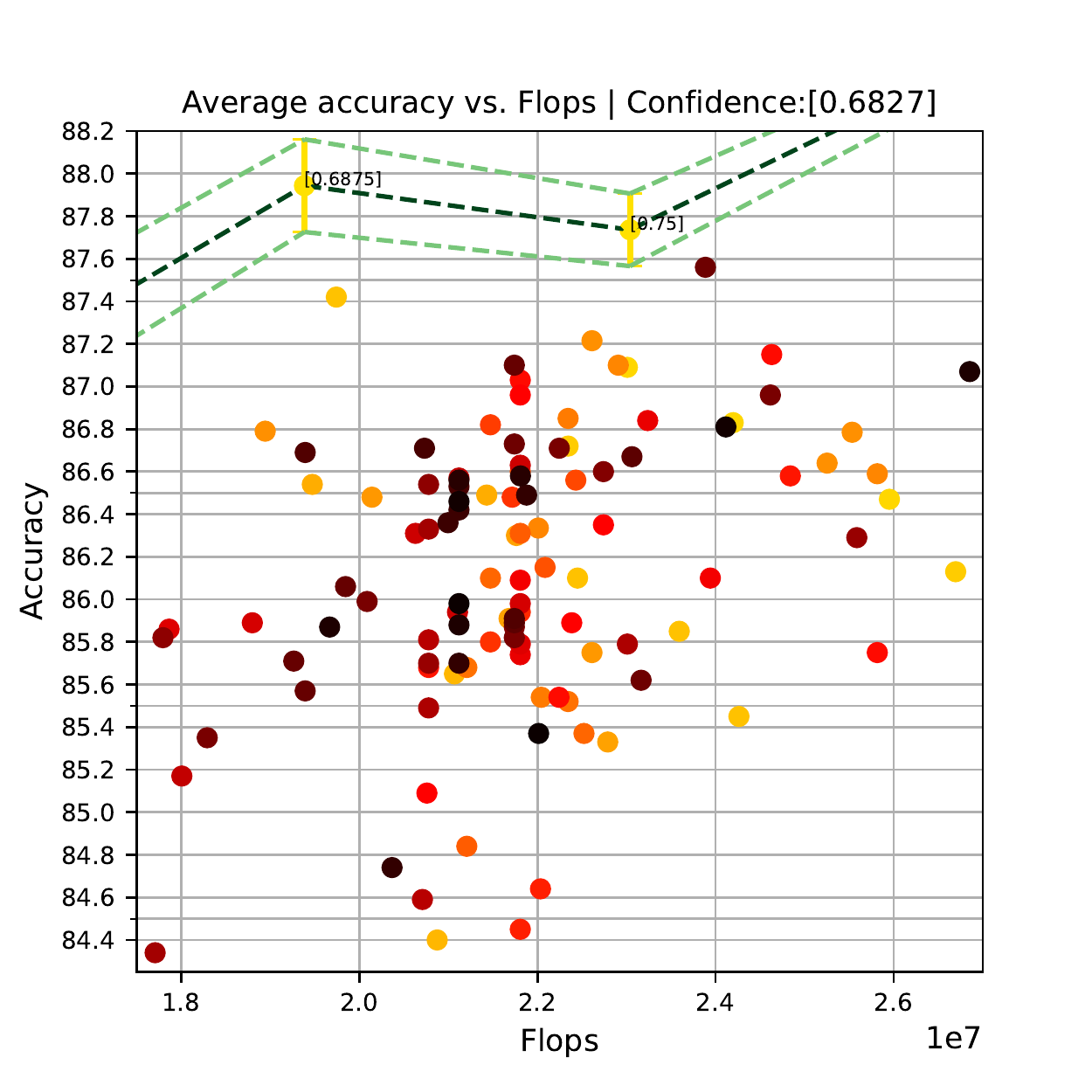}
        \subcaption{Learning rate = 1E-4. Initial distribution is 0.75, \ie $\opProbSingle=0.75$.}
    \end{subfigure}
  \caption{Results of search with interpolation loss.}
  \label{fig:pm4_results}
\end{figure}

\clearpage
\section{Search algorithms}

\subsection{Quantization search algorithm} \label{app:searchalgq}
\begin{algorithm}
\caption{Search method}
\label{alg:NICE_search_method}
\begin{algorithmic}[1]
    \STATE{Split training set $\trainSet$ to two halves:
    \begin{itemize}
        \item $\searchSet$, a set to update distribution parameters $\alpha$
        \item $\trainWeightsSet$, a set to update network weights $\networkWeights$.
    \end{itemize}
     }
    \STATE{Set bitwidth set $\layerOpsSet$ for each layer}
    \STATE{Set an initial distribution $\prob^{0}$}
    \STATE{Set a configurations subset size $|\configSubset{}|$ for $\J$ gradient estimator}
    \STATE{Set $\trainWeightsEpochs$, number of epochs to train network weights in each iteration}
    \STATE{Set a target homogeneous configuration $a_{homogeneous}$ for $\complexityLoss$}
    \STATE{Set $\lambda$ for $\complexityLoss$}
    \STATE{Set a function $\complexitySigma$ for $\complexityLoss$}
    \WHILE{not converged}
        \FOR{1 to $\trainWeightsEpochs$}
            \FOR{batch $b$ in $\trainWeightsSet$}
                \STATE{Sample configuration $\networkConfig$ from current distribution $\prob^{k}$}
                \STATE{Update $\networkWeights$ by a gradient step on $\ceLoss$}
            \ENDFOR
        \ENDFOR
        \FOR{batch $b$ in $\searchSet$}
            \STATE{Sample configurations subset $\configSubset{k}$ from current distribution $\prob^{k}$}
            \STATE{Update the distribution parameters $\alpha$ by a gradient step on $\J$}
        \ENDFOR
    \ENDWHILE
    \STATE{Sample and evaluate found configurations}
\end{algorithmic}
\end{algorithm}

\clearpage
\subsection{Basic pruning search method} \label{app:searchi}
\begin{algorithm}
\caption{Basic search method}
\label{alg:Slimmable_search_method_I}
\begin{algorithmic}[1]
    \STATE{Split training set $\trainSet$ to two halves:
    \begin{itemize}
        \item $\searchSet$, a set to update distribution parameters $\alpha$
        \item $\trainWeightsSet$, a set to update network weights $\networkWeights$.
    \end{itemize}
     }
    \STATE{Set an initial distribution $\prob^{0}$}
    \STATE{Set a homogeneous configurations set $A$ for Slimmable method weights training}
    \STATE{Set a configurations subset size, $|\configSubset{}|$, for $\J$ gradient estimator}
    \STATE{Set $\trainWeightsEpochs$, number of epochs to train network weights in each iteration}
    \STATE{Set a target homogeneous configuration $a_{homogeneous}$ for $\complexityLoss$}
    \STATE{Set $\lambda$ for $\complexityLoss$}
    \STATE{Set a function, $\complexitySigma$, for $\complexityLoss$}
    \WHILE{not converged}
        \STATE{Set Slimmable training method to train on $A \cup \expectedConfig{k}$}
        \FOR{1 to $\trainWeightsEpochs$}
            \FOR{batch $b$ in $\trainWeightsSet$}
                \STATE{Update $\networkWeights$ by Slimmable training method gradient step}
            \ENDFOR
        \ENDFOR
        \FOR{batch $b$ in $\searchSet$}
            \STATE{Sample configurations subset $\configSubset{k}$ from current distribution $\prob^{k}$}
            \STATE{Train each configuration $\networkConfig \in \configSubset{k}$ weights for 5 epochs over $\trainWeightsSet$}
            \STATE{Update the distribution parameters $\alpha$ by a gradient step on $\J$}
        \ENDFOR
    \ENDWHILE
    \STATE{Sample and evaluate found configurations}
\end{algorithmic}
\end{algorithm}

\clearpage
\subsection{Resetting the $\omega$} \label{app:searchii}
\begin{algorithm}
\caption{Resetting the $\omega$}
\label{alg:Slimmable_search_method_II}
\begin{algorithmic}[1]
    \STATE{Set training set $\trainSet$}
    \STATE{Set an initial distribution $\prob^{0}$}
    \STATE{Set a homogeneous configurations set $A$ for Slimmable method weights training}
    \STATE{Set a configurations subset size, $|\configSubset{}|$, for $\J$ gradient estimator}
    \STATE{Set $\trainWeightsEpochs$, number of epochs to train network weights in each iteration}
    \STATE{Set $\trainWeightsInterval$, number of iterations between $\networkWeights$ update}
    \STATE{Set a target homogeneous configuration $a_{homogeneous}$ for $\complexityLoss$}
    \STATE{Set $\lambda$ for $\complexityLoss$}
    \STATE{Set a function, $\complexitySigma$, for $\complexityLoss$}
    \WHILE{not converged}
        \IF{$ \left( \trainWeightsInterval \mod k \right) == 0$}
            \STATE{Set Slimmable training method to train on $A \cup \expectedConfig{k}$}
            \STATE{Set random values to $\networkWeights$}
            \FOR{1 to $\trainWeightsEpochs$}
                \FOR{batch $b$ in $\trainSet$}
                    \STATE{Update $\networkWeights$ by Slimmable training method gradient step}
                \ENDFOR
            \ENDFOR
        \ENDIF
        \STATE{Sample configurations subset $\configSubset{k}$ from current distribution $\prob^{k}$}
        \STATE{Train each configuration $\networkConfig \in \configSubset{k}$ weights for 5 epochs over $\trainSet$}
        \FOR{batch $b$ in $\trainSet$}
            \STATE{Update the distribution parameters $\alpha$ by a gradient step on $\J$}
        \ENDFOR
    \ENDWHILE
    \STATE{Sample and evaluate found configurations}
\end{algorithmic}
\end{algorithm}
In addition, we decided to make a few more changes:
\begin{itemize}
    \item To optimize runtime, we decided to perform the training $\networkWeights$, which takes most of the time, once in $10-20$ of updating the distribution parameters $\alpha$.
    \item We used the whole training set both for updating $\alpha$ and $\networkWeights$.
    \item We decided to save more running time by evaluated the configurations in $\configSubset{k}$ on each batch in $\trainSet$, instead of evaluation on a single batch, which further reduced runtime.
\end{itemize}

\clearpage
\subsection{Disabling weight-sharing} \label{app:searchiii}
\begin{algorithm}
\caption{Disabling weight-sharing}
\label{alg:Slimmable_search_method_III}
\begin{algorithmic}[1]
    \STATE{Set training set $\trainSet$}
    \STATE{Set an initial distribution $\prob^{0}$}
    \STATE{Set a configurations subset size, $|\configSubset{}|$, for $\J$ gradient estimator}
    \STATE{Set $\trainWeightsEpochs$, number of epochs to train a configuration weights $\configWeights$}
    \STATE{Set a target homogeneous configuration $a_{homogeneous}$ for $\complexityLoss$}
    \STATE{Set $\lambda$ for $\complexityLoss$}
    \STATE{Set a function, $\complexitySigma$, for $\complexityLoss$}
    \WHILE{not converged}
        \STATE{Sample configurations subset $\configSubset{k}$ from current distribution $\prob^{k}$}
        \FOR{configuration $\networkConfig \in \configSubset{k}$}
            \STATE{Set random values to $\configWeights$}
            \FOR{1 to $\trainWeightsEpochs$}
                \FOR{batch $b$ in $\trainSet$}
                    \STATE{Update $\configWeights$ by a gradient step on $\ceLossGeneric{\configIndices}$}
                \ENDFOR
            \ENDFOR
        \ENDFOR
        \FOR{batch $b$ in $\trainSet$}
            \STATE{Update the distribution parameters $\alpha$ by a gradient step on $\J$}
        \ENDFOR
    \ENDWHILE
    \STATE{Sample and evaluate found configurations}
\end{algorithmic}
\end{algorithm}

\clearpage
\subsection{Interpolation loss} \label{app:searchiv}

\begin{algorithm}
\caption{Interpolation loss}
\label{alg:Slimmable_search_method_IV}
\begin{algorithmic}[1]
    \STATE{Set training set $\trainSet$}
    \STATE{Set an initial distribution $\prob^{0}$}
    \STATE{Set a configurations subset size, $|\configSubset{}|$, for $\J$ gradient estimator}
    \STATE{Set $\trainWeightsEpochs$, number of epochs to train a configuration weights $\configWeights$}
    \STATE{Set a function, $\complexitySigma$, for $\lossConfig$}
    \WHILE{not converged}
        \STATE{Sample configurations subset $\configSubset{k}$ from current distribution $\prob^{k}$}
        \FOR{configuration $\networkConfig \in \configSubset{k}$}
            \STATE{Set random values to $\configWeights$}
            \FOR{1 to $\trainWeightsEpochs$}
                \FOR{batch $b$ in $\trainSet$}
                    \STATE{Update $\configWeights$ by a gradient step on $\ceLossGeneric{\configIndices}$}
                \ENDFOR
            \ENDFOR
        \ENDFOR
        \FOR{batch $b$ in $\trainSet$}
            \STATE{Update the distribution parameters $\alpha$ by a gradient step on $\J$}
        \ENDFOR
    \ENDWHILE
    \STATE{Sample and evaluate found configurations}
\end{algorithmic}
\end{algorithm}

\subsubsection{The expected loss}
The expected loss is calculated by the linear interpolation between any two consecutive homogeneous configurations. For a heterogeneous configuration as $\networkConfig$ with $\networkComplexity$ arithmetic complexity, let $\ceLossInterpolation{a}$ be an approximation of cross-entropy of homogeneous configuration of complexity $a$.
Then the loss of some configuration will be:
\begin{equation}
    \lossConfig = \sigma \left( \ceLoss{} - \ceLossInterpolation{\networkConfig} \right)
\end{equation}
for some increasing function $\complexitySigma$, \eg LeakyReLU, sigmoid or identity. Note there is no explicit arithmetic complexity loss term.
For an approximation, we used a linear interpolation between two homogeneous configurations, $\homogeneousConfig{1}$ and $\homogeneousConfig{2}$, with the closest arithmetic complexity to $\networkConfig$, such that 
\begin{equation}
    \homogeneousComplexity{1} \leq \networkComplexity \leq \homogeneousComplexity{2}
\end{equation}
 for a some predefined list of homogeneous configurations for which  the average loss over 5 different training sessions is calculated before training.
 
\clearpage
\section{Multinomial distribution lemmas}

\begin{lemma}
Let
\begin{itemize}
    \item Layer $\ell$ contains $\nFiltersLayer$ filters.
    \item $\layerOpsSet$ is a set of possible operations in layer $\ell$.
    \item $\layerRV{\ell}$ is a random variable from a multinomial distribution, \ie
    \begin{equation}
        \layerRV{\ell} \sim \multinomial \left( \nFiltersLayer,(\opProb{1},\opProb{2},...,\opProb{|\layerOpsSet|}) \right)
    \end{equation}
\end{itemize}
The probability to sample configuration $\layerConfigDetailed$ is:
\begin{equation}
    \Pr ( \layerRV{\ell} = \layerConfig{\ell} ) = \multinomialConfigProb
\end{equation}

\begin{proof}
\begin{equation}
\begin{split}
    \layerConfigProb &= \multinomialFrac \cdot \layerOpsProdK (\opProb{k})^{\opSample{k}} \\
    &= \multinomialFrac \cdot \layerOpsProdK \left( \opProbDetailed{k} \right) ^{\opSample{k}} \\
    &= \multinomialFrac \cdot \frac{\layerOpsProdK \opSampleNumeratorMultinomial}{\layerOpsProdK \left( \opSampleDenominatorMultinomial \right) ^{\opSample{k}}} \\
    &= \multinomialFrac \cdot \frac{\exp \left\{ \layerOpsSumK \opAlpha{k} \cdot \opSample{k} \right\} }{\left( \opSampleDenominatorMultinomial \right) ^{\layerOpsSumK \opSample{k}}} \\
    &= \multinomialConfigProb
\end{split}
\end{equation}
\end{proof}
\end{lemma}

\begin{lemma}
The partial derivative of the probability to sample layer configuration $\layerConfig{\ell}$ under the multinomial distribution is:
\begin{equation}
    \opAlphaDerivative \layerConfigProb = \multinomialConfigDerivative
\end{equation}

\begin{proof}
\begin{equation}
\begin{split}
     \opAlphaDerivative \layerConfigProb &= \opAlphaDerivative \Bigg( \multinomialConfigProb \Bigg) \\
    &= \multinomialFrac \cdot \opAlphaDerivative \multinomialConfigProbFrac \\
    &= \multinomialFrac \cdot \frac{1}{\opSampleDenominatorMultinomialWithPower \cdot \opSampleDenominatorMultinomialWithPower} \\
    & \Bigg[ \opSampleDenominatorMultinomialWithPower \cdot \opAlphaDerivative \opSampleNumeratorMultinomialSum - \\
    &- \opSampleNumeratorMultinomialSum \cdot \opAlphaDerivative \opSampleDenominatorMultinomialWithPower \Bigg] \\
    &= \multinomialFrac \cdot \Bigg( \opSample{t} \cdot \frac{\opSampleNumeratorMultinomialSum}{\opSampleDenominatorMultinomialWithPower} - \\
    &- \nFiltersLayer \cdot \frac{\exp \left\{ \opSample{t} \right\}}{\opSampleDenominatorMultinomial} \cdot \frac{\opSampleNumeratorMultinomialSum}{\opSampleDenominatorMultinomialWithPower} \Bigg) \\
    &= \opSample{t} \cdot \layerConfigProb - \nFiltersLayer \cdot \opProb{t} \cdot \layerConfigProb \\
    &= \multinomialConfigDerivative
\end{split}
\end{equation}
\end{proof}
\end{lemma}

\begin{lemma}
The partial derivative of the probability $\probConfig$ to sample network configuration $\networkConfig$ under the multinomial distribution is:
\begin{equation}
    \opAlphaDerivative \probConfig = \multinomialConfigProbDerivative{}
\end{equation}

\begin{proof}
\begin{equation}
\begin{split}
     \opAlphaDerivative \probConfig &= \opAlphaDerivative \prodLayersProb{r} \\
    &= \prodLayersProbLeaveOut{r} \cdot \opAlphaDerivative \layerConfigProb \\
    &= \prodLayersProbLeaveOut{r} \cdot \left( \multinomialConfigDerivative \right) \\
    &= \multinomialConfigDerivativeDifferencePart \cdot \prodLayersProb{r} \\
    &= \multinomialConfigProbDerivative{}
\end{split}
\end{equation}
\end{proof}
\end{lemma}

\begin{lemma} \label{multi_loss_pdv}
The partial derivative of the loss expected value $\J$ under the multinomial distribution is:
\begin{equation}
    \opAlphaDerivative \J = \multinomialLossEVderivative
\end{equation}

\begin{proof}
\begin{equation}
\begin{split}
     \opAlphaDerivative \J &= \opAlphaDerivative \configsSum \\
    &= \configsSumOperator \lossConfig \cdot \opAlphaDerivative \probConfig \\
    &= \multinomialLossEVderivative
\end{split}
\end{equation}
\end{proof}
\end{lemma}
where $\opSampleConfig{}$ represents on how many filters in layer $\ell$ in configuration $\networkConfig$ we apply operation $t$.

\section{Binomial distribution lemmas}

\begin{lemma}
Let
\begin{itemize}
    \item Layer $\ell$ contains $\nFiltersLayer$ filters.
    \item $\layerRV{\ell}$ is a random variable from a binomial distribution, \ie
    \begin{equation}
        \layerRV{\ell} \sim \binomial \left( \nFiltersLayer-1,\opProbSingle \right)
    \end{equation}
\end{itemize}
The probability to sample configuration $\layerConfigSingleDetailed$ is:
\begin{equation}
    \Pr \left( \layerRV{\ell} = \layerConfig{\ell} \right) = \binomialConfigProb
\end{equation}

\begin{proof}
\begin{equation}
\begin{split}
    \Pr \left( \layerRV{\ell} = \layerConfig{\ell} \right) &= \binom{\nFiltersLayer-1}{\opSampleSingle} \cdot \left( \opProbSingle \right) ^{\opSampleSingle} \cdot \left( 1-\opProbSingle \right) ^{\binomialComplementPower} \\
    &= \binomialFrac \cdot \opProbBinomialDetailedWithPower \cdot \left( 1 - \opProbBinomialDetailed \right) ^{\binomialComplementPower} \\
    &= \binomialFrac \cdot \opProbBinomialDetailedWithPower \cdot \left( \frac{1}{\opProbBinomialDetailedDenominator} \right) ^{\binomialComplementPower} \\
    &= \binomialConfigProb
\end{split}
\end{equation}
\end{proof}
\end{lemma}

\begin{lemma}
The partial derivative of the probability to sample layer configuration $\layerConfig{\ell}$ under the binomial distribution is:
\begin{equation}
    \opAlphaSingleDerivative \layerConfigProb = \binomialConfigDerivative
\end{equation}

\begin{proof}
\begin{equation}
\begin{split}
     \opAlphaSingleDerivative \layerConfigProb &= \opAlphaSingleDerivative \left( \binomialConfigProb \right) \\
    &= \binomialFrac \cdot \opAlphaSingleDerivative \binomialConfigProbFrac \\
    &= \binomialFrac \cdot \frac{1}{\opProbBinomialDetailedDenominatorWithPower \cdot \opProbBinomialDetailedDenominatorWithPower} \\
    & \Big( \opProbBinomialDetailedDenominatorWithPower \cdot \opAlphaSingleDerivative \opProbBinomialDetailedNumeratorWithPower \\
    &- \opProbBinomialDetailedNumeratorWithPower \cdot \opAlphaSingleDerivative \opProbBinomialDetailedDenominatorWithPower \Big) \\
    &= \binomialFrac \cdot \Bigg( \frac{\opSampleSingle \cdot \opProbBinomialDetailedNumeratorWithPower}{\opProbBinomialDetailedDenominatorWithPower} \\
    &- \nTrialsBinomial \cdot \binomialConfigProbFrac \cdot \frac{\opProbBinomialDetailedNumerator}{\opProbBinomialDetailedDenominator} \Bigg) \\
    &= \binomialFrac \cdot \binomialConfigProbFrac \cdot \binomialConfigDerivativeDifferencePart{} \\
    &= \binomialConfigDerivative
\end{split}
\end{equation}
\end{proof}
\end{lemma}

\begin{lemma}
The partial derivative of the probability $\probConfig$ to sample network configuration $\networkConfig$ under the binomial distribution is:
\begin{equation}
    \opAlphaDerivative \probConfig = \binomialConfigProbDerivative{}
\end{equation}

\begin{proof}
\begin{equation}
\begin{split}
     \opAlphaSingleDerivative \probConfig &= \opAlphaSingleDerivative \prodLayersProb{r} \\
    &= \prodLayersProbLeaveOut{r} \cdot \opAlphaSingleDerivative \layerConfigProb \\
    &= \prodLayersProbLeaveOut{r} \cdot \left( \binomialConfigDerivative \right) \\
    &= \binomialConfigDerivativeDifferencePart{} \cdot \prodLayersProb{r} \\
    &= \binomialConfigProbDerivative{}
\end{split}
\end{equation}
\end{proof}
\end{lemma}

\begin{lemma}  \label{bi_loss_pdv}
The partial derivative of the loss expected value $\J$ under the binomial distribution is:
\begin{equation}
    \opAlphaSingleDerivative \J = \binomialLossEVderivative
\end{equation}

\begin{proof}
\begin{equation}
\begin{split}
     \opAlphaSingleDerivative \J &= \opAlphaSingleDerivative \configsSum \\
    &= \configsSumOperator \lossConfig \cdot \opAlphaSingleDerivative \probConfig \\
    &= \binomialLossEVderivative
\end{split}
\end{equation}
\end{proof}
\end{lemma}
where $\opSampleSingleConfig{}$ represents on how many filters in layer $\ell$ in configuration $\networkConfig$ we apply the operation.

\section{BOPs definition and loss derivation} \label{app:bops}
Under quantization as a constraint, we use BOPs as arithmetic complexity metric. The BOPs metric quantifying the number of bit operations. Given the bitwidth of two operands, it is possible to approximate the number of bit operations required for a basic arithmetic operation such as addition and multiplication.

An important phenomenon is the non-linear relation between the number of activation and weight bits and the resulting network complexity in BOPs. To quantify this effect, let us consider a single convolutional layer with $b_\mathrm{w}$-bit weights and $b_\mathrm{a}$-bit activations containing $n$ input channels, $m$ output channels, and $k \times k$ filters. The maximum value of a single output is about $2^{b_\mathrm{a}+b_\mathrm{w}} n k^2$, which sets the accumulator width in the MAC operations to $b_\mathrm{o} = b_\mathrm{a}+b_\mathrm{w} + \log_2 {nk^2}$. The complexity of a single output calculation consists therefore of $nk^2$ $b_\mathrm{a}$-wide $\times$ $b_\mathrm{w}$-wide multiplications and about the same amount of $b_\mathrm{o}$-wide additions. This yields the total layer complexity of
    \begin{eqnarray}
    	\mathrm{BOPs} \, \approx \, mnk^2( b_\mathrm{a} b_\mathrm{w} + b_\mathrm{a}+b_\mathrm{w} + \log_2 {nk^2}).
    \end{eqnarray}
Note that the reduction of the weight and activation bitwidth decreases the number of BOPs as long as the factor $b_\mathrm{a} b_\mathrm{w}$ dominates the factor $\log_2 nk^2$. Since the latter factor depends only on the layer topology, this point of diminishing return is network architecture-dependent. Another factor that must be incorporated into the BOPs calculation is the cost of fetching the parameters from an an external memory. Two assumptions are made in the approximation of this cost: firstly, we assume that each parameter is only fetched once from an external memory; secondly, the cost of fetching a $b$-bit parameter is assumed to be $b$ BOPs. Given a neural network with $n$ parameters all represented in $b$ bits, the memory access cost is simply $nb$.

\subsection{BOPs loss derivation}  

\begin{figure*}[t]
\begin{center}
\includegraphics[width=.99\linewidth]{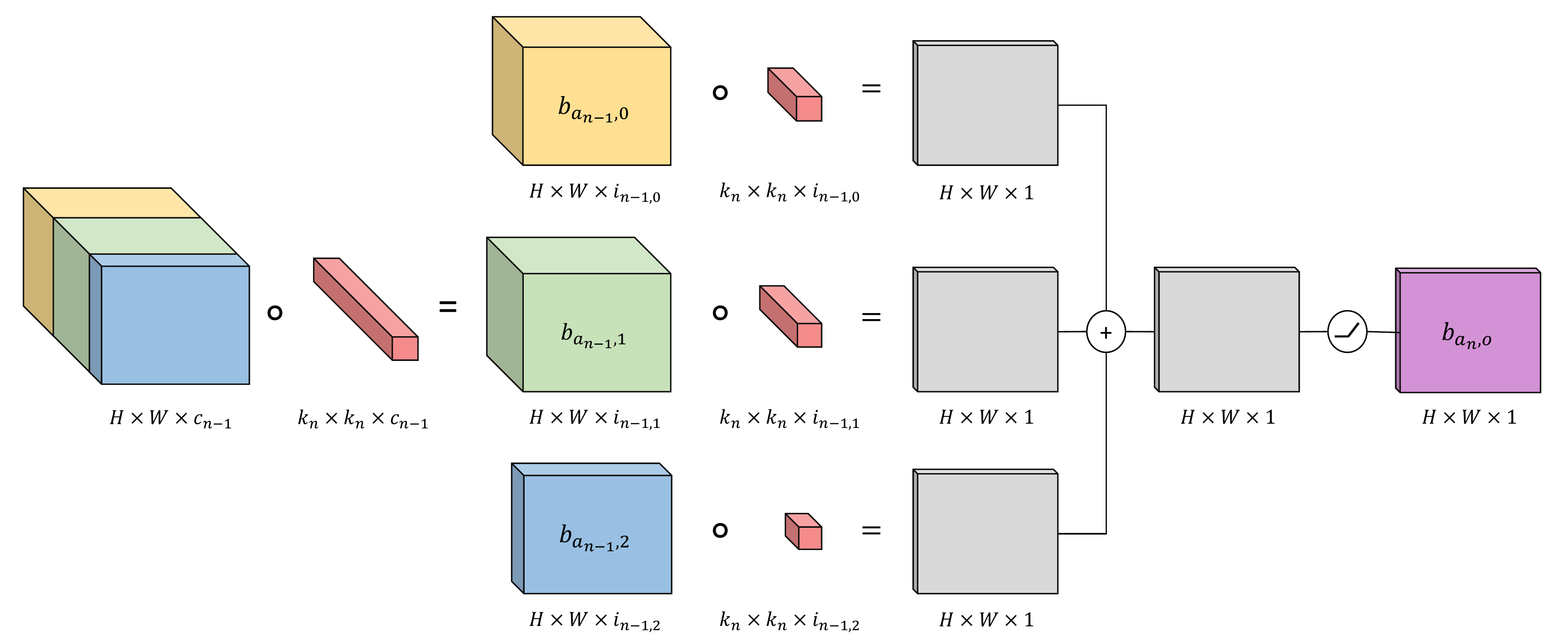}
\end{center}
\caption{Applying a $\left( b_{\omega_{n},o}, b_{a_{n},o} \right)$ filter of the $n$-th layer, \ie the filter weights are quantized using $b_{\omega_{n},o}$ bits and its activation (output) uses $b_{a_{n},o}$ bits.}
\label{fig:filter}
\end{figure*}

Since custom precision data types are used for the network weights and activations, 
the number of MAC operations \cite{kahan1996ieee}
is not an appropriate metric to acurately estimate the computational complexity of the desired model. 
Recently, Baskin~\etal~\cite{baskin2018uniq} 
proposed \emph{Bit OPerations (BOPs)} as a metric to quantify the computational complexity
of neural networks with multiple bitwidths.
However, since we need to take into account filter-wise granularity, the exact expression for BOPs is slightly different. 

%
%

Nevertheless, the main idea is the same. We denote by $b^{\omega}_{\oper} $  and $b^{\alpha}_{\oper}$  bitwidth of weights and activations for operation type $t$. 
For a filter in the $n$-th convolutional layer, the filter's output is of shape $c_{n} \times H \times W$.
For simplicity we assume the filter is a $k_n\times{k_n}$ square. 

The operation of convolution can be viewed as alternating multiplications of the input pixel by the weight and addition of the result to the accumulator, which stores the result of the convolution. While the cost of multiplication is obviously $b^{\alpha}_{\oper_1}b^{\omega}_{\oper_2} $
, the cost of addition is a bit harder to approximate.

The accumulator needs to be able to store any possible result of the convolution, and thus approximate the required bitwidth with maximal value of a single filter of bitwidth $b^{\omega}_{\oper_2} $, which is a sum of maximal values of each multiplication
\begin{equation}
\begin{split}
M_{n,\oper_2} 
= 
\sum_{\oper_{1} \in \operations}
\idx_{n-1,o_1} 2 ^ {b^{\omega}_{\oper_2} + b^{a}_{\oper_1}}
\tdu{k}{n}{2} 
=
2^{b^{\omega}_{\oper_2}}
\tdu{k}{n}{2}
\sum_{\oper_{1}=1}^m
\idx_{n-1,\oper_1} 2 ^ {b^{a}_{\oper_1}},
\end{split}
\end{equation}
%
which sets the accumulator width to

\begin{equation}
\begin{split}
b^{\text{AW}}_{n,\oper_2} 
=
\log_{2}{M_{n,\oper_2} }
\approx
b^{\omega}_{\oper_2} \log_{2}{
    \sum_{\oper_{1} \in \operations} \idx_{n-1,\oper_1} 2 ^ {b^{\alpha}_{o_1}}
}.
\end{split}
\end{equation} 

Therefore, 
the complexity of computing the single pixel of a single filter is
\begin{equation}
\bops_{n,\oper_2}\left(\vidx\right) = \tdu{k}{n}{2}\qty[c_{n-1}b^{\text{AW}}_{n,\oper_2}
+
\sum_{\oper_{1} \in \operations}
\idx_{n-1,\oper_1}
b^{\alpha}_{\oper_1}
b^{\omega}_{\oper_2} ]
\end{equation} 
This yields the total layer complexity of

\begin{equation} \label{eq:complexity}
\bops_n\left(\vidx\right)
\approx HW \sum_{\oper_{1} \in \operations} \bops_{n,\oper_2}\left(\vidx\right)
\end{equation} 


The proposed metric is useful when the inference is performed on a custom hardware like FPGAs or ASICs. 
Both are natural choices for quantized networks, 
due to the use of lookup tables (LUTs) and dedicated MAC (or more general DSP) units, 
which are efficient with custom data types.


\end{document}